\documentclass{article}

\PassOptionsToPackage{numbers,sort&compress}{natbib}
\usepackage[main, preprint]{neurips_2026}

\makeatletter
\renewcommand{\@noticestring}{Accepted at the 40th Conference on Neural Information Processing Systems (NeurIPS 2026).}
\makeatother

\usepackage[utf8]{inputenc}
\usepackage[T1]{fontenc}
\usepackage{courier}
\usepackage{xurl}
\usepackage{booktabs}
\usepackage{array}
\usepackage{graphicx}
\usepackage{amsfonts}
\usepackage{amsmath}
\usepackage{amssymb}
\usepackage{amsthm}
\usepackage{mathtools}
\usepackage{nicefrac}
\usepackage{microtype}
\usepackage{xcolor}
\usepackage{placeins}
\usepackage{needspace}
\usepackage{algorithm}
\usepackage{algpseudocode}
\usepackage{hyperref}

\hypersetup{
  colorlinks=true,
  citecolor=blue!50!black,
  linkcolor=blue!50!black,
  urlcolor=blue!50!black
}

\newcommand{\E}{\mathbb{E}}

\newcommand{\diag}{\operatorname{diag}}
\newcommand{\CVaR}{\operatorname{CVaR}}

\newcommand{\KL}{\operatorname{KL}}
\newcommand{\argmax}{\operatorname*{arg\,max}}
\newcommand{\op}{\mathrm{op}}
\newcommand{\FG}{\mathrm{FG}}
\newcommand{\res}{\mathrm{res}}

\newcommand{\rid}{\mathrm{rid}}
\newcommand{\eff}{\mathrm{eff}}
\newcommand{\TTI}{\mathrm{TTI}}
\newcommand{\loss}{\ell}

\newtheorem{theorem}{Theorem}
\newtheorem{proposition}{Proposition}
\newtheorem{appendixproposition}{Proposition}[section]

\title{Fisher-Glass Learning: Classification Bounds and Subpopulation Audits}

\author{%
  El Mustapha Mansouri \\
  School of Engineering, Institute of Science Tokyo \\
  Yokohama, Kanagawa 226-8501, Japan \\
  \texttt{mansouri.e.2224@m.isct.ac.jp}
}

\begin{document}

\maketitle

\begin{abstract}
Accurate classifiers can fail when background cues change without altering the correct labels. Predictive performance alone does not quantify how much task information remains after accounting for such nuisance variation. Here we extend the Fisher-glass framework to connect nuisance-adjusted information with local classification difficulty and subpopulation auditing. We derive lower bounds on local minimax excess classification risk under explicit distinguishability and decision-separation conditions, alongside statistical audits accounting for estimation error and regularization. On Waterbirds, a feature-based diagnostic of task--background alignment selects an accurately classified group, with selection frozen before held-out evaluation. Across five training seeds, background replacements preserving the extracted bird foreground and labels yield 93.2\% accuracy when background type is preserved and 40.6\% when it changes, a controlled difference of 52.6 percentage points. Separately, search using task labels and a fixed nuisance direction identifies an independently evaluated population with 98.8\% membership-weighted accuracy and low estimated projected predictive Fisher information. Training and acquisition experiments show that information gains and predictive gains can diverge. These findings support joint assessment of predictive performance, information under declared nuisance models and controlled sensitivity when diagnosing vulnerability and evaluating model improvements.
\end{abstract}

\section{Introduction}
\label{sec:introduction}

\begin{figure}[t]
\centering
\includegraphics[width=\linewidth]{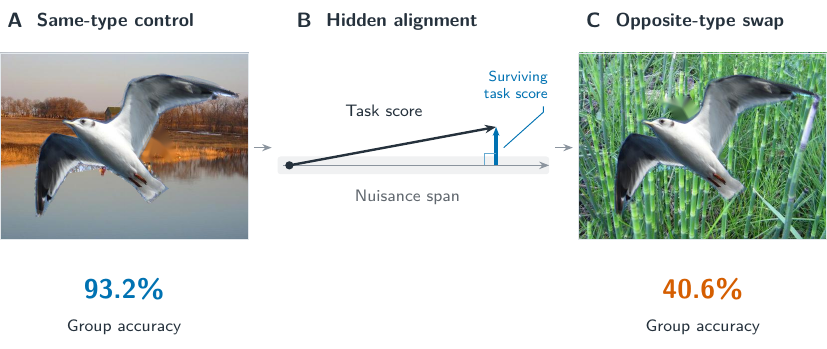}
\caption[Background sensitivity despite high control accuracy]{\input{figures/caption01_main.tex}}
\label{fig:concept}
\end{figure}

A classifier can predict observed labels accurately while the data weakly distinguish a specified task effect from admissible nuisance variation. This ambiguity matters for classification when statistically similar alternatives prescribe different labels. Autonomous bird monitoring provides a practical motivation: species classifiers operate under variable illumination and occlusion \citep{mansouri2025autonomous}. We ask when it creates a learning obstacle, and how to audit the remaining task information on predefined and learned populations. Group-robust learning and error-based reweighting emphasize subpopulation performance \citep{sagawa2019distributionally,liu2021just}; our objective is to establish which information claims the data support and assess their relevance through controlled nuisance interventions (Figure~\ref{fig:concept}).

Efficient-score projection removes task variation that nuisance scores can reproduce \citep{bickel1993efficient,tsiatis2006semiparametric}. The Fisher-glass framework of \mbox{\citet{mansouri_arai_2026_fisher_glasses}} connects this geometry to inverse-information tails and a population certificate. We develop its classification consequence: decision separation turns limited distinguishability into excess-risk bounds. A worked regression family shows that constant raw information at a reference parameter can conceal an effective sample size set by the mass near the decision boundary. Statistical corrections then determine which population information statements survive estimation error, regularization, and adaptive selection.

\begin{samepage}
On Waterbirds, a frozen feature-based rule selects a group with 92.0\% source accuracy and a 52.6-point controlled drop between same-type and opposite-type background replacements. The information, accuracy, and negative log-likelihood (NLL) selectors identify the same eligible group, whose controlled accuracy drop quantifies background sensitivity. Within that group, nuisance-derived coordinates improve species-held-out prediction of intervention failures beyond a specified confidence baseline. A separate search discovers an accurate population with low predictive information; its sensitivity to interventions remains untested.\par
\end{samepage}

Our contributions are:
\begin{itemize}
\item \textbf{A classification consequence.} We derive local excess-risk bounds with explicit decision separation and radius, and exhibit a classification family with matching minimax rates and an interpretable loss of effective sample size.
\item \textbf{Population information audits.} We account for estimation error, ridge bias, membership normalization, and support correction, and evaluate frozen learned slices independently. The audit distinguishes supported information floors from uncertain or adaptively selected estimates.
\item \textbf{Behavioral evaluation.} Controlled interventions and confidence-baseline comparisons test the relevance of nuisance-derived measurements. Training and acquisition evaluate whether information gains correspond to predictive gains, which need not occur together.
\end{itemize}

\section{Task--Nuisance Geometry and Information Collapse}
\label{sec:geometry}

Let $e$ index an observed group or latent subpopulation, and let $\Pi$ be the declared distribution over these environments. The audit fixes the task contrast, admissible nuisance variations, and common task-coordinate normalization. Information describes the task distinction remaining after nuisance adjustment.

\subsection{Task Information After Nuisance Profiling}
\label{sec:task-information}

Consider a local model $P_{\theta,\mu,e}$ with task parameter $\theta$ and nuisance parameter $\mu$. A likelihood score measures the local change in log likelihood as a parameter varies. At $P_e=P_{0,0,e}$, assume differentiability in quadratic mean, with centered, square-integrable task-score vector $s_{T,e}$ and nuisance-score vector $s_{N,e}$. Let $\mathcal T_{N,e}$ be the closed span of admissible nuisance scores in $L^2(P_e)$, with inner product $\langle f,g\rangle_e=\E_{P_e}[fg]$, and let $\Pi_{\mathcal T_{N,e}}$ denote orthogonal projection onto that span. The \textbf{efficient task score} and its information are
\begin{equation}
\begin{aligned}
s^{\eff}_{T,e}&=s_{T,e}-\Pi_{\mathcal T_{N,e}}s_{T,e},\\
F_e:=F^{\eff}_{T\mid N,e}
&=\E_{P_e}[s^{\eff}_{T,e}(s^{\eff}_{T,e})^\top]
=J_{TT,e}-J_{TN,e}J_{NN,e}^{-1}J_{NT,e}.
\end{aligned}
\label{eq:efficient-score-information}
\end{equation}
The projection acts componentwise. The Schur expression applies when finite nuisance scores span $\mathcal T_{N,e}$ and $J_{NN,e}\succ0$, with $J_{AB,e}=\E_{P_e}[s_{A,e}s_{B,e}^{\top}]$ \citep{bickel1993efficient,tsiatis2006semiparametric,cox1987parameter}. Write $\lambda_e=\lambda_{\min}(F_e)$ for the weakest task information. A zero eigenvalue identifies a task direction contained in the nuisance span.

For scalar task and nuisance scores with positive raw information, their Fisher correlation $\rho_e=J_{TN,e}/\sqrt{J_{TT,e}J_{NN,e}}$ gives
\begin{equation}
F_e=J_{TT,e}(1-\rho_e^2),
\qquad
\underbrace{F_e/J_{TT,e}}_{\text{surviving-information fraction}}=1-\rho_e^2.
\label{eq:surviving-information-fraction}
\end{equation}
Absolute information measures remaining curvature in the declared coordinates. The fraction measures the proportion retained, lies in $[0,1]$, and is invariant to nonzero scalar rescaling of either score. It compares alignment across slices with different raw information; the prospective selection statistic uses a ridge-regularized feature--Schur analogue with fixed coordinate scales. The Fisher-Glass certificate summarizes absolute information across environments.

\subsection{Task--Nuisance Alignment and Loss--Information Separation}
\label{sec:tangency-separation}

Let $b_e,r_e$ be centered, orthogonal, unit-norm scores, with nuisance span $\operatorname{span}\{b_e\}$. For $\delta_e\ge0$, define the task score $a_e=b_e+\delta_e r_e$. Then
\begin{equation}
a_e=b_e+\delta_e r_e,
\quad J_{TT,e}=1+\delta_e^2,
\quad F_e=\delta_e^2,
\quad \frac{F_e}{J_{TT,e}}=\frac{\delta_e^2}{1+\delta_e^2}.
\label{eq:tangency-identities}
\end{equation}
Projection removes $b_e$, leaving $\delta_e r_e$; raw information stays at least one as efficient information vanishes. Section~\ref{sec:synthetic-validation} illustrates low observed loss alongside the corresponding feature-moment geometry. A classification cost additionally requires different optimal decisions, which the next section makes explicit. The inherited tail mechanism is illustrated in Appendix Figure~\ref{fig:geometry-tails}.

\section{Classification Difficulty and Population Information}
\label{sec:tail-identifiability}

We study the statistical consequences of nuisance projection at two scales. Within an environment, local classification bounds and a worked family connect weak task information to learning difficulty under explicit distinguishability and decision-separation conditions. Across environments, geometry and probability mass near Fisher-null sets govern inverse-information tails, motivating a certificate of population information coverage.

\subsection{Efficient Information and Local Classification Difficulty}
\label{sec:classification-difficulty}

Parameter ambiguity need not imply classification difficulty: probabilities $0.90$ and $0.95$ prescribe the same label. A lower bound requires statistically similar alternatives whose optimal decisions disagree on enough probability mass. Define classification risk $R_P(g)=P\{g(X)\ne Y\}$, Bayes risk $R_P^\star=\inf_g R_P(g)$, and excess risk $\mathcal E_P(g)=R_P(g)-R_P^\star$. For a specified local family, let $\mathfrak R_e(n)$ denote minimax expected excess classification risk based on $n$ independent observations from environment $e$. Write $\KL(P\Vert Q)$ for the Kullback--Leibler divergence.

\begin{theorem}[Local classification difficulty]
\label{thm:local-classification-difficulty}
Fix $\lambda_e>0$ and a unit weakest efficient-information direction $u_e$. For every $0<t\le t_{0,e}$, suppose the local family contains $P_{e,+t},P_{e,-t}$ with task coordinates $\pm tu_e$ and nuisance coordinates allowed to vary. Assume, for constants $K_e,c_e,t_{0,e}>0$ and $q>0$,
\begin{equation}
\KL(P_{e,+t}\Vert P_{e,-t})\le K_et^2\lambda_e,
\qquad
\inf_g\{\mathcal E_{P_{e,+t}}(g)+\mathcal E_{P_{e,-t}}(g)\}\ge c_et^q.
\label{eq:local-classification-assumptions}
\end{equation}
Then, for a universal constant $c_0>0$,
\begin{equation}
\mathfrak R_e(n)\ge c_0c_e
\min\{t_{0,e}^q,(nK_e\lambda_e)^{-q/2}\}.
\label{eq:local-classification-lower-bound}
\end{equation}
\end{theorem}

The KL bound limits distinguishability; decision separation supplies the classification cost. Both must hold over radius $t_{0,e}$. With quadratic separation, the interior rate is $c_e/(nK_e\lambda_e)$, capped by the local decision scale. Appendix~\ref{app:theory-classification} proves the testing reduction; the next example calculates both conditions and a matching upper bound.

\begin{samepage}
\subsection{A Worked Classification Mechanism}
\label{sec:classification-mechanism}
Let $X\sim\operatorname{Unif}[-1,1]$, $0<\delta\le1/2$, $|\theta|,|\mu|\le\delta/8$, and
\begin{equation}
\eta_{\theta,\mu}(x):=P_{\theta,\mu}(Y=1\mid X=x)
=\tfrac12+\tfrac{x}{4}+\theta+\mu\mathbf1\{|x|>\delta\}.
\label{eq:boundary-region-family}
\end{equation}
Here $\delta$ is boundary-region half-width and probability mass. The nuisance reproduces a task shift outside that region; the Bayes rule is $g_\theta^\star(x)=\mathbf1\{x\ge-4\theta\}$ throughout the family. At $(\theta,\mu)=(0,0)$,
\begin{equation}
J_{TT}=4\log3,
\qquad F_\delta=4\log\frac{2+\delta}{2-\delta}\asymp\delta.
\label{eq:boundary-region-information}
\end{equation}
The alternatives $(t,-t)$ and $(-t,t)$ agree outside $|X|\le\delta$, have decision separation $4t^2$, and satisfy $\KL(P_{+t}\Vert P_{-t})\le5F_\delta t^2$ for $0<t\le\delta/8$. Theorem~\ref{thm:local-classification-difficulty} and a matching estimator give minimax risk over this parameter rectangle, uniformly in $n\ge1$ and $\delta$,
\begin{equation}
\mathfrak R_\delta(n)\asymp\min\{\delta^2,(n\delta)^{-1}\}.
\label{eq:boundary-region-minimax-rate}
\end{equation}
Approximately $n\delta$ observations distinguish the task effect from nuisance variation where the decision boundary can move. The $\delta^2$ cap retains the shrinking local decision scale. This statistical nuisance preserves Bayes labels at fixed $\theta$; it does not model the physical background-swap operation. Appendix~\ref{app:classification-mechanism} gives the calculation and a finite-sample upper bound.\par
\end{samepage}

\subsection{Information Tails Across Populations}
\label{sec:collapse-geometry}

Population aggregation describes how often weak task information occurs; classification still retains the radius and separation factors above. The inherited geometric law is:
\begin{theorem}[Geometric tail law {\citep{mansouri_arai_2026_fisher_glasses}}]
\label{thm:inverse-information-geometry}
Under the uniform regularity and localization assumptions in Appendix~\ref{app:theory-tails}, let the Fisher-null set have codimension $c$, information vanish to order $2m$, and nearby density have order $\nu>-c$. Then, for some $C>0$,
\begin{equation}
\Pr(\lambda(z)^{-1}>x)\sim Cx^{-\beta},
\qquad \beta=\frac{c+\nu}{2m}.
\label{eq:inverse-information-tail-exponent}
\end{equation}
\end{theorem}
Quadratic collapse with positive density gives $\beta=c/2$; the inverse mean is finite exactly when $\beta>1$. Appendix~\ref{app:theory-tails} contains the calculation and Figure~\ref{fig:geometry-tails}.

\begin{samepage}
\subsection{The Fisher-Glass Certificate}
\label{sec:fg-certificate}

Upper-tail conditional value-at-risk, $\CVaR_\alpha$, averages the largest $1-\alpha$ fraction of a quantity, with fractional weighting at atoms \citep{rockafellar2000optimization,rockafellar2002conditional}. We apply the Fisher-glass certificate \citep{mansouri_arai_2026_fisher_glasses} to classification environments $e\sim\Pi$, using the weakest task information $\lambda_e$ and stabilization $\varepsilon$:
\begin{equation}
L_e^{(\varepsilon)}=(\lambda_e+\varepsilon)^{-1},
\qquad
I_\alpha^{\FG}=\bigl[\CVaR_{\alpha,e\sim\Pi}(L_e^{(\varepsilon)})\bigr]^{-1},
\quad 0<\alpha<1.
\label{eq:fisher-glass-certificate}
\end{equation}
\end{samepage}
A larger certificate means a smaller inverse-information tail, expressed in information units. Stabilization $\varepsilon>0$ caps inverse information at $1/\varepsilon$; it does not remove the underlying null geometry. Under Theorem~\ref{thm:inverse-information-geometry}'s power-law regime, the limit $\varepsilon\downarrow0$ is positive for $\beta>1$ and vanishes for $\beta\le1$.

\begin{proposition}[Information coverage]
\label{prop:information-coverage}
For $\rho>\varepsilon$,
\begin{equation}
I_\alpha^{\FG}\ge\rho
\quad\Longrightarrow\quad
\Pr_{e\sim\Pi}(\lambda_e\ge\rho-\varepsilon)\ge\alpha.
\label{eq:information-coverage}
\end{equation}
\end{proposition}

The certificate establishes an absolute information floor on an $\alpha$-fraction of the declared environment distribution; classification consequences retain the local radius and decision separation. Appendix~\ref{app:theory-certificate} gives the proof. With four equally weighted groups and $\alpha=0.95$, it reduces to the minimum stabilized information, so that real-data audit does not isolate a benefit from tail aggregation.

\section{Computing and Auditing Fisher-Glass}
\label{sec:computing-auditing}

The audit establishes which population information statements are supported after accounting for how the matrix and population were constructed. It fixes the task contrast, nuisance hypothesis, matrix, normalization, and audit distribution before independent evaluation. Ridge can raise an information estimate, adaptive selection can exaggerate extremes, and weak support can destabilize weighted populations. The corrections below address these distinct sources of distortion; fitted nuisance probes and temperatures are frozen before final evaluation.

\subsection{Operational Scores and Nuisance Coordinates}
\label{sec:operational-scores}

For a fitted binary predictive probability $p_w(x)$, including any fixed temperature calibration, introduce task and nuisance directions $z(x)$ and $b(x)$ through $\operatorname{logit}p_{\theta,\mu}(x)=\operatorname{logit}p_w(x)+\theta z(x)+\mu b(x)$. At $\theta=\mu=0$, the score is $(Y_i-p_i)v_i$, with $v_i=(z_i,b_i)^\top$ and $p_i=p_w(x_i)$. Feature moments use $M_i=v_iv_i^\top$; predictive Fisher uses $M_i=p_i(1-p_i)v_iv_i^\top$. The prospective Waterbirds audit uses the uncentered feature--Schur fraction of task-logit and background-probe coordinates. Hidden-slice discovery uses temperature-calibrated predictive Fisher, mixing 90\% of the gated matrix with 10\% of the current split's same-class mean matrix before projection, during both search and held-out evaluation. This corresponds to shrinkage $\eta=0.10$; feature moments supply a comparison. Appendix~\ref{app:operational-matrices} specifies these constructions, the observed-label score alternative, operational subspaces, and computational cost.

For the scalar, unregularized, unshrunk predictive target with positive nuisance information,
\begin{equation}
F_{\mathrm{pred}}=\min_a\E[p(X)(1-p(X))\{z(X)-a b(X)\}^2].
\label{eq:predictive-confidence-decomposition}
\end{equation}
Small information can therefore reflect task--nuisance alignment or confident probabilities with small $p(1-p)$. Neither mechanism alone establishes intervention sensitivity. Nuisance coordinates come from metadata, fitted annotation probes, or a fixed semantic direction; without annotations, that direction is the hypothesis being audited. Appendices~\ref{app:hidden-slice-discovery} and~\ref{app:predefined-diagnostics} report stability and specification checks.

\subsection{Ridge Estimation and Lower Information Bounds}
\label{sec:stable-estimation}

Let $M_i$ be the per-example task--nuisance matrix under the chosen construction, with task and nuisance blocks indexed by $T$ and $N$. For a predefined slice $g$ of size $n_g$, form $\widehat J_g=n_g^{-1}\sum_{i\in g}M_i$ and compute the ridge Schur estimate
\begin{equation}
\widehat F_g^{\rid}
=\widehat J_{TT,g}-\widehat J_{TN,g}
(\widehat J_{NN,g}+\lambda_{\rid}I)^{-1}\widehat J_{NT,g}.
\label{eq:ridge-estimator}
\end{equation}
Real-data computations use $\lambda_{\rid}=10^{-3}$. Ridge stabilizes residualization but raises information relative to the unregularized target. Let $a_g$ bound estimation error and $d_g$ bound the ridge-to-unregularized gap on a simultaneous event over the audited slices. For declared weights $\Pi_G$, use
\begin{equation}
\begin{aligned}
\underline\lambda_g&=[\lambda_{\min}(\widehat F_g^{\rid})-a_g-d_g]_+,\\
\underline I_{\alpha,G}&=
\left[\CVaR_{\alpha,g\sim\Pi_G}
\left((\underline\lambda_g+\varepsilon)^{-1}\right)\right]^{-1}.
\end{aligned}
\label{eq:lower-certificate}
\end{equation}
Here $[x]_+=\max\{x,0\}$. On that event, this lower-bounds the unregularized population certificate. A large lower bound supports a positive information floor; a small lower bound alone does not establish small true information. Certification also requires a positive lower confidence bound for $\lambda_{\min}(J_{NN,g})$. The real-data audits use Bonferroni-adjusted Gaussian/Wald moment simulation at nominal 95\% confidence, conditional on fitted models, slices, and normalization. These implemented intervals differ from the concentration guarantees under Appendix~\ref{app:estimation-certification}'s assumptions.

\subsection{Hidden-Slice Discovery and Held-Out Certification}
\label{sec:hidden-slice-method}

A gate $h_\phi(x,y)\in[0,1]$ assigns membership weights. Discovery maximizes inverse projected information subject to effective-sample-size, maximum-weight, and class-mass constraints. We normalize memberships within class, cap excessive weights, and mix with uniform weights as needed to satisfy the effective-sample-size constraint. Five validation folds assess transfer and stability; a final gate is fitted on all discovery data, frozen, and evaluated on an untouched split. The support correction is recomputed on evaluation data under the frozen gate. Its unshrunk target is the weighted ratio $J_\phi=\E[h_\phi M]/\E[h_\phi]$, with positive membership mass. Uncertainty must account for the numerator, membership normalization, and adjustments used to enforce the weight constraints; the Waterbirds target retains the stated shrinkage. Appendix~\ref{app:hidden-slice-discovery} specifies the constraints, objective, and held-out target.

\begin{algorithm}[H]
\caption{Fisher-Glass audit}
\label{alg:fisher-glass-audit}
\begin{algorithmic}[1]
\State \textbf{Specify:} predictor, matrix, coordinates, normalization, calibration, and audit weights.
\State Fix predefined slices, or fit and freeze a constrained gate on discovery data.
\State \textbf{Audit independently:} check support and conditioning; compute ridge estimates, unregularized estimates, and lower confidence bounds.
\State \textbf{Report:} raw and projected information, surviving fraction, slice support, tail certificate, and uncertainty.
\end{algorithmic}
\end{algorithm}

\section{Evaluating Information Audits}
\label{sec:experiments}

The experiments test distinct claims: sensitivity of a frozen predefined group, incremental failure prediction beyond confidence measures, the inherited geometric tail law, and held-out discovery of a declared information target. The predefined-group and learned-slice studies use different models and matrices.

\subsection{Prospective Diagnosis Under a Nuisance Intervention}
\label{sec:prospective-diagnosis}

On Waterbirds \citep{sagawa2019distributionally}, the four groups cross bird class with background type and are written \emph{class/background}. We fully fine-tune ImageNet-initialized MobileNetV3-small models \citep{howard2019searching} with empirical risk minimization (ERM). Validation data fit the temperature and cross-fitted background probe. Eligible slices have accuracy at least the overall validation accuracy and negative log-likelihood (NLL) at most the median across slices. Among them, we select the smallest surviving fraction, $\lambda_{\min}(\widehat F_g^{\rid})/\widehat J_{TT,g}$. After pilot experiments, this rule is fixed for five additional training seeds, 22--26. Each checkpoint, temperature, probe, selected slice, and audit constant is frozen before intervention scoring.

The rule selects \textbf{waterbird/water in all five seeds}: its validation surviving fraction is $24.8\%$, compared with $59.0\%$ for landbird/land, the other eligible slice. Held-out interventions recompose birds onto same-type or opposite-type backgrounds using the same extracted foreground layer, matte, and task labels. Same-type recomposition controls for the image-construction procedure. Selected-slice accuracy is $92.0\%$ on original images, $93.2\%$ under the control, and $40.6\%$ under opposite-type replacement: a \textbf{$52.6\pm1.4$ percentage-point controlled drop}, with a nominal 95\% normal-approximation interval across seeds.

\begin{figure}[t]
\centering
\includegraphics[width=\linewidth]{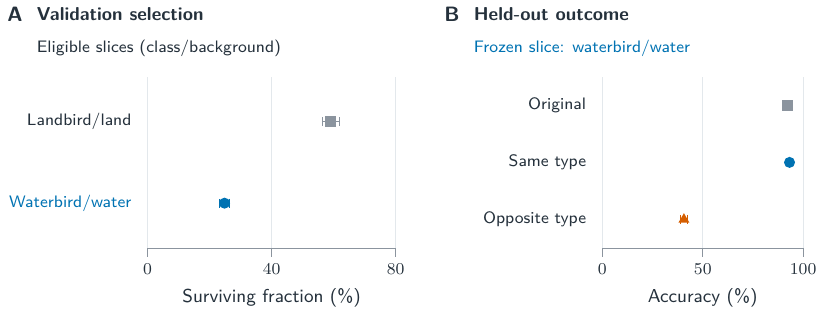}
\caption[A 52.6-point controlled accuracy drop in the selected group]{\input{figures/caption_prospective_main.tex}}
\label{fig:prospective-main}
\end{figure}

Minimum validation accuracy and maximum validation NLL select the same eligible group in all five seeds (Figure~\ref{fig:prospective-main}). Among eligible slices, minimum raw and absolute projected information select landbird/land, with a $26.7$-point drop. Centering switches selection in all five seeds, identifying slice means as a ranking contributor; predictive Fisher retains waterbird/water in all five. Appendix~\ref{app:prospective-intervention} reports matched selectors, matrix and aggregation comparisons, and individual flip prediction.

\paragraph{Predicting failures beyond confidence.}
Within the selected group, a retained auxiliary analysis predicts opposite-type failures among originally correct birds using five-fold cross-fitting by species. The baseline includes source NLL, calibrated true-label confidence, absolute raw logit, and standardized task logit $z$. Adding the background component $\beta_g b$ and residual magnitude $|z-\beta_g b|$ improves average precision by $0.0257$ (nominal 95\% paired interval $[0.0211,0.0302]$), area under the receiver operating characteristic curve (AUROC) by $0.0150$, and mean log likelihood by $0.0128$ nats per donor row. All gains are positive in all five seeds; intervals use $1.96$ standard errors of paired seed differences (Table~\ref{tab:d-flip-prediction}). This establishes incremental value of the added coordinates. It does not isolate the residual's contribution from the background score or test the scalar population certificate as an individual predictor.

\FloatBarrier

\subsection{Synthetic Classification and Tail Checks}
\label{sec:synthetic-validation}

Draw independent $s,r\sim N(0,1)$, set $z=s+\delta_e r$ and $Y=\mathbf1\{z>0\}$, and evaluate the nuisance-only logit $4s$. Environments use $\delta_e=0.35U_e^{1/\gamma}$, $U_e\sim\mathrm{Uniform}(0,1)$, with $\gamma=1$ in the main experiment. Five-seed mean accuracy is $94.54\%$ and $\CVaR_{0.95}$ of environment-mean NLL is $0.262$. The information audit uses exact population feature moments of $(z,s)$, giving $J_{TT,e}=1+\delta_e^2$ and $\lambda_e=\delta_e^2$. Its certificate is $I^{\FG}_{0.95}=(1.22\pm0.53)\times10^{-8}$, with $\varepsilon=10^{-12}$.

The inverse-information exponent is $0.498\pm0.003$, matching $\beta=\gamma/2=0.5$ in the geometric tail law of Theorem~\ref{thm:inverse-information-geometry} (Figure~\ref{fig:geometry-tails}). Log-survival regression uses the largest $5\%$ of inverse-information observations. Additional experiments check $\beta=K/2$ for $K$ independent information branches (Appendix~\ref{app:theory-tails}) and the information floor obtained by allocating observations to an informative source (Appendix~\ref{app:information-reserves}). Uncertainties are nominal 95\% normal-approximation intervals across five seeds. These checks exhibit favorable prediction alongside weak projected information and the predicted tail behavior (Appendix~\ref{app:synthetic-checks}).

\subsection{Discovering Low-Information Subpopulations}
\label{sec:low-information-subpopulations}

In a separate experiment, we fit a linear task head on frozen ImageNet MobileNetV3-small representations. Gate discovery uses task labels and a nuisance score obtained by projecting CLIP image embeddings onto a fixed water-minus-land text direction \citep{radford2021learning}. Background and group annotations remain sealed through test evaluation. We search each task class separately with a class-specific logistic gate and select the class--gate pair with highest discovery utility. Gates target $10\%$ within-class mass under Section~\ref{sec:hidden-slice-method}'s support constraints. The validation split supplies discovery data: five folds assess transfer, after which the gate is refitted on the full split, frozen, and evaluated once on untouched test data.

\begin{figure}[t]
\centering
\includegraphics[width=\linewidth]{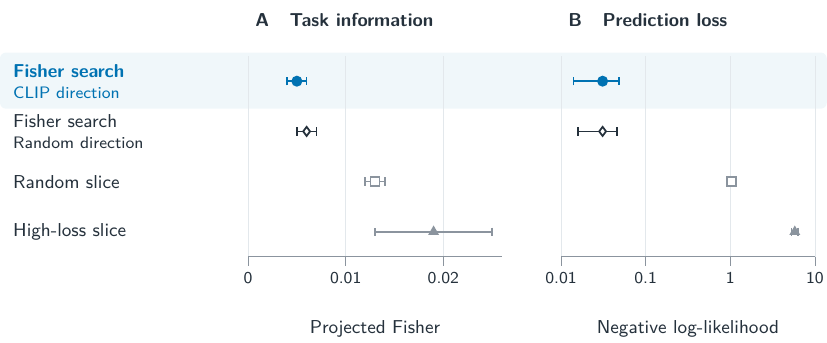}
\caption[Discovery finds accurate slices with low projected predictive information]{\input{figures/caption03_main.tex}}
\label{fig:discovery}
\end{figure}

The discovered slice has \textbf{$98.8\%$ membership-weighted accuracy} and lower estimated projected predictive Fisher information than high-loss and random selectors targeting the same mass in every seed (Figure~\ref{fig:discovery}). Its calibrated predictive information is $0.005\pm0.001$, and its uncalibrated NLL is $0.031\pm0.017$, with 95\% Student-$t$ intervals across five paired seeds. Accuracy and NLL use support-corrected membership weights without information pooling (Appendix~\ref{app:reported-slice-metrics}). The Fisher and high-loss gates' top $10\%$ raw-membership sets within their selected class have mean Jaccard overlap $0.001$, indicating largely distinct selections.

The random-direction control attains similar predictive information, with the paired CLIP-minus-random interval spanning zero. Feature-moment comparisons with high-loss selection also have paired intervals containing zero. Predictive Fisher includes the confidence factor $p(1-p)$, whereas displayed NLL is uncalibrated; these results do not separate confidence from nuisance alignment. They establish discovery of a low-information population, not its unusual intervention sensitivity or superiority to confidence-matched selection. Appendix~\ref{app:discovery-results} gives paired intervals, support diagnostics, and composition.

\section{Information-Guided Interventions}
\label{sec:improving-information}

We evaluate whether information changes improve the intended predictive outcome. An informative observation source can guarantee an information floor under the Fisher-reserve design \citep{mansouri_arai_2026_fisher_glasses} (Appendix~\ref{app:information-reserves}); training and acquisition require separate behavioral evaluation. For predictor parameters $w$, empirical task-risk objective $\widehat R_{\mathrm{base}}$ (ERM or group distributionally robust optimization, Group DRO), and penalty $\lambda_{\FG}$, training uses equally weighted groups:
\begin{equation}
\min_w\;\widehat R_{\mathrm{base}}(w)
+\lambda_{\FG}\operatorname{CVaR}_{\alpha,g}
\!\left[\bigl\{\lambda_{\min}(\widehat F_g^{\rid}(w))+\varepsilon\bigr\}^{-1}\right].
\label{eq:fg-training-compact}
\end{equation}
Coordinates follow Section~\ref{sec:computing-auditing}. Among the frozen-feature models whose validation worst-group accuracy is within two percentage points of the best, we select the coefficient with the largest validation certificate. This selection uses validation group labels.

\begin{table}[htbp]
\centering
\caption{Frozen-feature Waterbirds training. Means with nominal 95\% normal-approximation intervals ($\pm1.96$ standard errors) across five seeds; uncentered feature--Schur ridge certificates use $\alpha=0.95$. Validation selection follows Section~\ref{sec:improving-information}.}
\label{tab:frozen-training}
\small
\begin{tabular}{@{}lccc@{}}
\toprule
Objective & \shortstack{Average\\accuracy (\%)} & \shortstack{Worst-group\\accuracy (\%)} & \shortstack{Fisher-Glass\\certificate} \\
\midrule
ERM & $77.2\pm0.1$ & $55.7\pm0.3$ & $0.198\pm0.001$ \\
ERM + Fisher-Glass & $80.4\pm0.2$ & $56.3\pm0.3$ & $0.399\pm0.006$ \\
Group DRO & $80.9\pm0.3$ & $67.3\pm0.3$ & $0.233\pm0.003$ \\
Group DRO + Fisher-Glass & $80.6\pm0.9$ & $69.4\pm1.3$ & $0.614\pm0.059$ \\
\bottomrule
\end{tabular}
\end{table}

Fisher-Glass approximately doubles ERM's certificate; with Group DRO it raises both the certificate and mean worst-group accuracy (Table~\ref{tab:frozen-training}). Stronger penalties can reduce accuracy (Appendix~\ref{app:extended-training}). These source-performance gains do not establish a repair of background sensitivity.

\FloatBarrier

In prospective fine-tuning, the validation certificate increases from $0.289$ to $0.330$, while controlled swap drops remain $52.9$ points for FG and $52.6$ for ERM. A separate Fisher-based frozen-slice repair under validation-accuracy constraints selects the unchanged model in all five seeds. On another cohort, post hoc selection raises rediscovered-slice information but reduces worst-group accuracy by $12.6$ points; selected coefficients match validation-NLL minimization in every seed (Appendix~\ref{app:extended-training}). Thus information improvement and the intended behavioral improvement are separate outcomes.

For candidate $a$ with modified feature/input $x_a^{\mathrm{cf}}$ and label $y_a$, \emph{tail-transverse influence} $\TTI(a)=G_\alpha(a)\loss_w(x_a^{\mathrm{cf}},y_a)$ weights current predictive loss by the first-order inverse-information CVaR decrease $G_\alpha(a)$ along $a$'s allocation path. This differentiates the Schur complement along its weakest task direction with tail-sensitive weights (Appendix~\ref{app:tail-transverse-influence}). Waterbirds ranks candidates by $\widetilde{\TTI}_\alpha(a)=\widetilde G_\alpha(a)\loss_w(x_a^{\mathrm{cf}},y_a)$, using the tail-weighted squared-residual surrogate $\widetilde G_\alpha$ defined in Appendix~\ref{app:feature-acquisition}.

Waterbirds candidates translate features by the within-class difference of background-group means, scaled by $\gamma_{\mathrm{cf}}$. Training data fit means and the nuisance probe; validation supplies candidates for retraining, and evaluation is held out. Budgets are fractions of original training size. At $\gamma_{\mathrm{cf}}=1$ and a $1\%$ budget, $\widetilde{\TTI}_\alpha$ yields a mean worst-group accuracy gain of $4.52$ points versus $3.27$ for high-loss selection; high loss produces the larger mean certificate gain. At this strength, $\widetilde{\TTI}_\alpha$ has the larger mean worst-group accuracy gain at budgets of $1\%$, $2\%$, and $5\%$; selection by $\widetilde G_\alpha$ alone yields smaller improvements (Appendix~\ref{app:feature-acquisition}).

\section{Related Work}
\label{sec:related-work}

\paragraph{Efficient information and classification.}
Efficient-score theory defines task information after projection away from nuisance tangent spaces \citep{bickel1993efficient,newey1990semiparametric,tsiatis2006semiparametric}. Fisher-glass theory connects nuisance-adjusted information to inverse-information tails in metrology \citep{mansouri_arai_2026_fisher_glasses}. Our analysis develops classification bounds and statistical guarantees for population audits. The classification argument uses local testing reductions \citep{lecam1986asymptotic,tsybakov2009introduction} under an explicit Bayes-decision separation condition. The operational analysis addresses Schur estimation, ridge effects, and independent evaluation of learned population gates.

\paragraph{Tail risk, robust learning, and gradient geometry.}
CVaR concentrates objectives on adverse outcomes \citep{rockafellar2000optimization,rockafellar2002conditional}; Group DRO applies robust optimization to predictive loss across groups \citep{sagawa2019distributionally}. Invariant Risk Minimization seeks shared optimal classifiers \citep{arjovsky2019invariant}, while Fishr regularizes domain-level gradient variances \citep{rame2022fishr}. We use inverse projected-information tail risk as a training penalty alongside empirical risk or Group DRO, and evaluate information and predictive performance together.

\paragraph{Hidden stratification and slice discovery.}
Hidden-stratification research motivates finer population evaluation \citep{oakdenrayner2020hidden}. GEORGE clusters representations to infer subclasses \citep{sohoni2020no}, JTT upweights initially misclassified examples \citep{liu2021just}, and AGRO learns adversarial group assignments using prediction loss \citep{paranjape2022agro}. Our search targets low predictive projected information given task labels and a fixed nuisance direction. Support constraints and an independently evaluated frozen gate make accurately classified, low-loss populations eligible audit targets.

\paragraph{Information-based design and acquisition.}
E-optimal design emphasizes the smallest information eigenvalue \citep{pukelsheim2006optimal}; BAIT uses Fisher information for neural active learning \citep{ash2021gone}. TTI measures local information gain along a weak nuisance-adjusted task direction, weights environments by tail sensitivity, and multiplies by current task loss. Acquisition experiments assess realized information and accuracy changes after retraining.

\section{Limitations}
\label{sec:limitations}

\paragraph{Specification and interpretation.}
Audits quantify information for a declared task contrast, nuisance space, representation, and population; admissible nuisance changes must preserve the task. Unregularized nuisance projection of predictive Fisher gives efficient information for the declared predictive model, while feature--Schur quantities describe residual feature geometry. Their relation to population efficient information depends on the score model. Coordinate scaling and matrix construction affect the audit; standardization does not establish representation invariance. The incremental contribution of semantic nuisance projection remains unresolved (Section~\ref{sec:experiments}). A general procedure for admitting nuisance specifications through relevance, task preservation, stability, and conditioning checks remains to be evaluated (Appendix~\ref{app:operational-matrices}).

\paragraph{Theory and uncertainty.}
The local classification bound requires distinguishability and quantitative Bayes-decision separation throughout its radius. The worked family verifies these conditions and a matching rate, but its statistical nuisance is not a model of physical background swaps. Coverage concerns the declared environment distribution and frozen population rules. Small mass and weak nuisance conditioning amplify uncertainty; ridge stabilization can inflate information, so positive ridge values alone do not establish identifiability (Section~\ref{sec:stable-estimation}; Appendix~\ref{app:ridge-unregularized}). Implemented Gaussian/Wald coverage is nominal and conditional on fitted models, evaluated families, and normalization; it omits uncertainty from evaluation-fitted normalization. Repeated use of certification data requires additional control (Section~\ref{sec:computing-auditing}; Appendix~\ref{app:estimation-certification}).

\paragraph{Empirical scope and interventions.}
Evidence centers on Waterbirds, supplementary CivilComments diagnostics, and synthetic constructions. The prospective rule uses fresh seeds of the development benchmark; transfer requires further evaluation. The auxiliary failure-prediction gain combines background and residual coordinates, without a background-only ablation. No retained experiment tests whether the learned low-information slice is more sensitive than a confidence-matched population. Information gains and coefficient-selection policies require separate behavioral validation (Section~\ref{sec:improving-information}). Reserve guarantees assume an informative anchor; acquisition uses group-informed feature shifts whose transfer to realizable task-preserving inputs remains open (Appendix~\ref{app:information-interventions}).

\section{Conclusion}
\label{sec:conclusion}

Task--nuisance ambiguity has a classification cost when statistically similar alternatives require different decisions. Our local bound makes that condition explicit, and a worked family shows how nuisance confounding reduces the effective sample size near a decision boundary despite constant raw information at the reference parameter. A matching upper bound identifies the learning rate and the cap imposed by the local decision scale. Population audits then account for estimation error, ridge bias, adaptive membership, and support, establishing which information statements are supported by data.

The retained Waterbirds experiments establish a 52.6-point controlled accuracy drop in a frozen predefined group and incremental species-held-out failure-prediction value from nuisance-derived coordinates beyond a specified confidence baseline. Independent evaluation validates a discovered low predictive-information population. Training and acquisition show that information gains and predictive gains can diverge.

The classification result, population audit, and behavioral evaluations have complementary roles. They connect ambiguity to learning difficulty, quantify a declared information target with uncertainty, and test its relationship to observed behavior. This supports evaluating information objectives together with predictive performance and controlled nuisance sensitivity when assessing a classifier or a proposed improvement.

\FloatBarrier
\label{end:main}

\begin{ack}
This work received no funding. The author declares no competing interests.
\end{ack}

\clearpage
\bibliographystyle{unsrtnat}
\begingroup
\renewcommand{\ttdefault}{cmtt}
\setlength{\bibsep}{4pt plus 0.5pt minus 0.5pt}
\interlinepenalty=10000
\bibliography{references_core,references_camera_ready}
\endgroup

\clearpage
\appendix
\raggedbottom
\makeatletter
\setlength{\@fptop}{0pt}
\makeatother
\numberwithin{equation}{section}
\numberwithin{figure}{section}
\numberwithin{table}{section}
\renewcommand{\theHfigure}{supp.\thesection.\arabic{figure}}
\renewcommand{\theHtable}{supp.\thesection.\arabic{table}}

\section{Proofs of the Core Results}
\label{app:core-proofs}
Throughout, task coordinates use the common normalization of Section~\ref{sec:geometry}, and environments follow the declared distribution $\Pi$.

\subsection{Projection, Normalization, and Task--Nuisance Tangency}
\label{app:theory-projection}

Use the centered, square-integrable scores and blocks $J_{AB,e}$ of Section~\ref{sec:task-information}. Suppose the $d_N$ coordinates of the finite nuisance-score vector span $\mathcal T_{N,e}$ and $J_{NN,e}\succ0$. With $\Gamma_e=J_{TN,e}J_{NN,e}^{-1}$,
\begin{equation}
\begin{aligned}
\mathbb E_{P_e}[(s_{T,e}-\Gamma_es_{N,e})s_{N,e}^{\top}]&=J_{TN,e}-\Gamma_eJ_{NN,e}=0,\\
F_e=\mathbb E_{P_e}[s_{T,e}^{\eff}(s_{T,e}^{\eff})^{\top}]&=J_{TT,e}-J_{TN,e}J_{NN,e}^{-1}J_{NT,e}.
\end{aligned}
\label{eq:app-efficient-schur}
\end{equation}
Because $\Gamma_es_{N,e}$ lies componentwise in the nuisance span, orthogonality identifies the residual as the efficient score in Equation~\eqref{eq:efficient-score-information} \citep{bickel1993efficient,tsiatis2006semiparametric}. Equivalently,
\begin{equation}
u^{\top}F_eu=\min_{v\in\mathbb R^{d_N}}\mathbb E_{P_e}[(u^{\top}s_{T,e}-v^{\top}s_{N,e})^2],
\qquad v^\star=J_{NN,e}^{-1}J_{NT,e}u.
\label{eq:app-variational-schur}
\end{equation}
Hence $F_e\succeq0$, with $u^{\top}F_eu=0$ exactly when $u^{\top}s_{T,e}$ belongs to the nuisance span. The projection definition extends directly to the closed span of an infinite-dimensional nuisance tangent space.

For separate invertible transformations $\widetilde s_T=Us_T$ and $\widetilde s_N=Vs_N$, the nuisance span is unchanged and
\begin{equation}
\widetilde s_T^{\eff}=Us_T^{\eff},\qquad \widetilde F=UFU^{\top};
\qquad \frac{F}{J_{TT}}=1-\rho^2\quad\text{in the scalar case}.
\label{eq:app-coordinate-congruence}
\end{equation}
Here $\rho=J_{TN}/\sqrt{J_{TT}J_{NN}}$ requires positive raw task and nuisance information. Exact information is invariant to nuisance-basis changes; task congruence preserves rank but generally changes eigenvalues. The scalar surviving fraction is invariant to nonzero task rescaling and invertible nuisance-basis changes. Nuisance ridge preserves the latter invariance only if $\lambda_{\rid}I$ becomes $\lambda_{\rid}VV^\top$ under $s_N\mapsto Vs_N$ (Appendix~\ref{app:estimation-certification}).

For the tangency construction in Section~\ref{sec:tangency-separation}, let $b_e,r_e\in L_0^2(P_e)$ be orthogonal unit vectors, $\mathcal T_{N,e}=\operatorname{span}\{b_e\}$, and $a_e=b_e+\delta_er_e$ with $\delta_e\ge0$. The projection coefficient is $\langle a_e,b_e\rangle_e/\|b_e\|_e^2=1$, so
\begin{equation}
a_e^{\eff}=\delta_er_e,\qquad F_e=\|\delta_er_e\|_e^2=\delta_e^2,
\qquad J_{TT,e}=1+\delta_e^2,\qquad \frac{F_e}{J_{TT,e}}=\frac{\delta_e^2}{1+\delta_e^2}.
\label{eq:app-tangency-identities}
\end{equation}
This establishes Equation~\eqref{eq:tangency-identities}; coexistence with low prediction risk additionally requires a successful nuisance-based predictor.

\subsection{Normal Coordinates and the Inverse-Information Tail}
\label{app:theory-tails}

The full assumptions of Theorem~\ref{thm:inverse-information-geometry} are as follows. The zero set $\mathcal Z=\{\lambda=0\}$ is a nonempty compact $C^2$ embedded manifold without boundary, of codimension $c\in\{1,\ldots,d\}$. Write $U_r(\mathcal Z)=\{z:\operatorname{dist}(z,\mathcal Z)<r\}$ for its tube, $N_s\mathcal Z$ for the normal space at $s$, $S(N_s\mathcal Z)$ for its unit sphere, and $\operatorname{vol}_{\mathcal Z}$ for induced volume. The nonnegative $\lambda$ is continuous on $\overline U_{r_\star}(\mathcal Z)$ and bounded away from zero outside it. In coordinates $z=s+r\omega$, assume uniformly over $\omega\in S(N_s\mathcal Z)$ and $s\in\mathcal Z$
\begin{equation}
\lambda(s+r\omega)=a(s,\omega)r^{2m}(1+o(1)),\qquad
p(s+r\omega)=b(s,\omega)r^\nu(1+o(1)),
\label{eq:null-set-expansions}
\end{equation}
where $p$ is the Lebesgue density, $m>0$, $\nu>-c$, and continuous $a,b$ have positive uniform lower and upper bounds. Put
\begin{equation}
\beta=\frac{c+\nu}{2m},\qquad
C_{\mathcal Z}=\frac1{c+\nu}\int_{\mathcal Z}\int_{S(N_s\mathcal Z)}
b(s,\omega)a(s,\omega)^{-\beta}\,d\sigma_s(\omega)\,d\operatorname{vol}_{\mathcal Z}(s).
\label{eq:app-tail-constant}
\end{equation}
The sphere measure $\sigma_s$ is unnormalized; for $c=1$, each of the two normal directions has mass one.

\begin{proof}[Proof of Theorem~\ref{thm:inverse-information-geometry}]
For any $0<r_0<r_\star$, continuity and absence of zeros on the compact annulus $r_0\le\operatorname{dist}(z,\mathcal Z)\le r_\star$ give a positive minimum of $\lambda$. Together with the bound outside the outer tube, this localizes $\{\lambda<t\}$ to $U_{r_0}(\mathcal Z)$ for all sufficiently small $t$.
The tubular volume element is $dz=j(s,r\omega)r^{c-1}\,dr\,d\sigma_s\,d\operatorname{vol}_{\mathcal Z}$, where the coordinate Jacobian satisfies $j=1+O(r)$ uniformly. Thus, for any $0<\zeta<1$, uniformity permits $r_0$ small enough that
\begin{equation}
\begin{aligned}
(1-\zeta)ar^{2m}&\le\lambda(s+r\omega)\le(1+\zeta)ar^{2m},\\
(1-\zeta)br^\nu&\le p(s+r\omega)j(s,r\omega)\le(1+\zeta)br^\nu
\end{aligned}
\qquad(0<r<r_0).
\label{eq:app-uniform-tube-bounds}
\end{equation}
For small $t$, all radial thresholds lie inside this tube. The event contains $r<(t/((1+\zeta)a))^{1/(2m)}$ and is contained in $r<(t/((1-\zeta)a))^{1/(2m)}$. Integrating uses $\int_0^{(t/a)^{1/(2m)}}br^{c+\nu-1}\,dr=ba^{-\beta}t^\beta/(c+\nu)$ and gives
\begin{equation}
(1-\zeta)(1+\zeta)^{-\beta}C_{\mathcal Z}
\le\frac{\Pr(\lambda<t)}{t^\beta}
\le(1+\zeta)(1-\zeta)^{-\beta}C_{\mathcal Z}.
\label{eq:app-small-information-sandwich}
\end{equation}
Let $t\downarrow0$ and then $\zeta\downarrow0$. Compactness and the coefficient bounds ensure $0<C_{\mathcal Z}<\infty$, so $\Pr(\lambda<t)\sim C_{\mathcal Z}t^\beta$. Since $c\ge1$, $\mathcal Z$ has zero Lebesgue and environment measure; substituting $t=x^{-1}$ proves the inverse tail in the next display.
\end{proof}

The tail-integral identity then yields, for every $p_0>0$,
\begin{equation}
\begin{aligned}
\Pr(\lambda^{-1}>x)&\sim C_{\mathcal Z}x^{-\beta},\\
\mathbb E[\lambda^{-p_0}]&=p_0\int_0^\infty x^{p_0-1}\Pr(\lambda^{-1}>x)\,dx<\infty
\quad\Longleftrightarrow\quad p_0<\beta.
\end{aligned}
\label{eq:app-inverse-moments}
\end{equation}
The bounded-$x$ part is finite; the tail has the convergence criterion of $\int^\infty x^{p_0-\beta-1}\,dx$. In particular, the inverse mean is finite exactly for $\beta>1$, with logarithmic divergence at $\beta=1$.

\paragraph{Scalar branches.}
For $\lambda=\delta^2$, $\Pr(\delta<t)\sim C_\delta t^\gamma$ gives exponent $\gamma/2$. More generally, let $\lambda_K=\sum_{k=1}^K w_k\delta_k^2$, $w_k>0$, where $\delta\in[0,\infty)^K$ has joint density $f$ continuous and positive at the origin from within the orthant. Changing variables gives the scalar portfolio law \mbox{\citep{mansouri_arai_2026_fisher_glasses}}:
\begin{equation}
\begin{aligned}
\Pr(\lambda_K<t)
&=\frac{t^{K/2}}{\sqrt{\prod_k w_k}}\int_{B_K^+}
 f\bigl(\sqrt{t/w_1}u_1,\ldots,\sqrt{t/w_K}u_K\bigr)\,du\\
&\sim\frac{f(0)v_K}{2^K\sqrt{\prod_k w_k}}\,t^{K/2}.
\end{aligned}
\label{eq:app-independent-branches}
\end{equation}
Here $B_K^+$ is the positive-orthant unit ball and $v_K$ the full ball's volume. Independence gives $f(0)=\prod_k f_k(0+)$ for branch marginal densities $f_k$; the inverse mean is finite for $K\ge3$. This calculation concerns additive scalar information or a fixed task direction. A minimum-eigenvalue conclusion instead requires applying Theorem~\ref{thm:inverse-information-geometry} to that eigenvalue.

\begin{figure}[t]
\centering
\includegraphics[width=\linewidth]{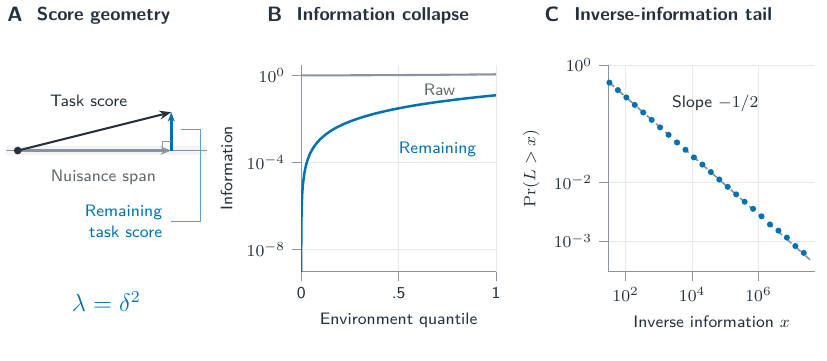}
\caption[Near-alignment produces a heavy inverse-information tail]{\input{figures/caption02_main.tex}}
\label{fig:geometry-tails}
\end{figure}

\subsection{Local Classification: Testing Reduction and Tail Aggregation}
\label{app:theory-classification}

Let $\mathcal E_P(g)=R_P(g)-R_P^\star$. Fix a local family $\mathcal P_e^{\mathrm{loc}}$ independently of sample size, and define
\begin{equation}
\mathfrak R_e(n)=\inf_{\widehat g_n}\sup_{P\in\mathcal P_e^{\mathrm{loc}}}
\mathbb E_{S_n\sim P^{\otimes n}}[\mathcal E_P(\widehat g_n(S_n))],
\label{eq:app-local-minimax}
\end{equation}
where learning rules are measurable and $n$ counts independent observations from environment $e$; independent learner randomization is included in the expectation. Assume $\lambda_e>0$ and a unit weakest efficient-information direction $u_e$. For every $0<t\le t_{0,e}$, the fixed family contains $P_{e,\pm t}$ with task coordinates $\pm tu_e$ and potentially varying nuisance coordinates. For fixed $K_e,c_e,t_{0,e}>0$ and $q>0$, assume both bounds in Equation~\eqref{eq:local-classification-assumptions} hold throughout that radius.

\begin{proof}[Proof of Theorem~\ref{thm:local-classification-difficulty}]
Fix an admissible $t$ and abbreviate $P_\pm=P_{e,\pm t}$. For any learning rule, the decision-separation condition $\mathcal E_{P_+}(g)+\mathcal E_{P_-}(g)\ge c_et^q$ implies that choosing the alternative with smaller excess risk makes any testing error cost at least $c_et^q/2$ under the true alternative. Explicitly,
\begin{equation}
\begin{aligned}
\psi_t(S_n)=+&\ \Longleftrightarrow\ 
\mathcal E_{P_+}(\widehat g_n(S_n))\le\mathcal E_{P_-}(\widehat g_n(S_n)),\\
\max_{\sigma\in\{+,-\}}\mathbb E_{P_\sigma^{\otimes n}}[\mathcal E_{P_\sigma}(\widehat g_n)]
&\ge\frac{c_et^q}{4}\{P_+^{\otimes n}(\psi_t=-)+P_-^{\otimes n}(\psi_t=+)\}.
\end{aligned}
\label{eq:app-risk-to-testing}
\end{equation}
The standard exponential two-point testing inequality \citep[Lemma~2.6]{tsybakov2009introduction}, together with product additivity of KL and the assumed KL bound, gives
\begin{equation}
\begin{aligned}
P_+^{\otimes n}(\psi_t=-)+P_-^{\otimes n}(\psi_t=+)
&\ge\tfrac12\exp[-n\operatorname{KL}(P_+\Vert P_-)]
\ge\tfrac12 e^{-nK_et^2\lambda_e},\\
\mathfrak R_e(n)&\ge\frac{c_e}{8}t^q e^{-nK_et^2\lambda_e}.
\end{aligned}
\label{eq:app-testing-reduction}
\end{equation}
The second line takes the supremum over the fixed family containing both alternatives and then the infimum over learning rules. Choose
\begin{equation}
\begin{aligned}
t_n&=\min\{t_{0,e},(nK_e\lambda_e)^{-1/2}\},\qquad nK_e\lambda_et_n^2\le1,\\
\mathfrak R_e(n)&\ge c_0c_e\min\{t_{0,e}^{q},(nK_e\lambda_e)^{-q/2}\},
\qquad c_0=\frac{e^{-1}}8.
\end{aligned}
\label{eq:app-local-risk-bound}
\end{equation}
Independent learner randomization has the same law under both alternatives and leaves their KL divergence unchanged, so the argument also covers randomized rules.
\end{proof}

\paragraph{Quadratic decision separation.}
Suppose the binary alternatives share a feature marginal $Q_e$, with regression functions $\eta_{\pm,t}(x)=P_{e,\pm t}(Y=1\mid X=x)$ and Bayes rules $g_{\pm,t}^\star(x)=\mathbf1\{\eta_{\pm,t}(x)\ge1/2\}$. Let $D_{e,t}=\{g_{+,t}^\star\ne g_{-,t}^\star\}$ be their disagreement set. Applying the binary excess-risk identity \citep[Eq.~(1.1)]{audibert2007fast} gives
\begin{equation}
\begin{aligned}
&\inf_g\{\mathcal E_{P_{e,+t}}(g)+\mathcal E_{P_{e,-t}}(g)\}\\
&\qquad=\int_{D_{e,t}}\min\{|2\eta_{+,t}(x)-1|,|2\eta_{-,t}(x)-1|\}\,dQ_e(x).
\end{aligned}
\label{eq:app-boundary-separation}
\end{equation}
Pointwise, a classifier must disagree with at least one Bayes rule on $D_{e,t}$; choosing the smaller margin attains the integral for unrestricted measurable binary classifiers, while it remains a lower bound for a restricted class. If, for every $0<t\le t_{0,e}$, a measurable $A_{e,t}\subseteq D_{e,t}$ has $Q_e(A_{e,t})\ge c_{\mathrm{mass},e}t$ and both margins at least $c_{\mathrm{margin},e}t$, with positive constants independent of $t$, the integral is at least $c_{\mathrm{mass},e}c_{\mathrm{margin},e}t^2$. Thus $q=2$ and $c_e=c_{\mathrm{mass},e}c_{\mathrm{margin},e}$ satisfy decision separation; the KL condition is still required separately.

\paragraph{Environmental aggregation.}
If these assumptions hold for $\Pi$-almost every environment and all quantities are measurable, monotonicity and positive homogeneity of CVaR give
\begin{equation}
\begin{aligned}
\mathcal B_e(n)&=c_e\min\{t_{0,e}^{q},(nK_e\lambda_e)^{-q/2}\},\\
\operatorname{CVaR}_{\alpha,e\sim\Pi}(\mathfrak R_e(n))
&\ge c_0\operatorname{CVaR}_{\alpha,e\sim\Pi}(\mathcal B_e(n)).
\end{aligned}
\label{eq:app-tail-risk-aggregation}
\end{equation}
Separation $c_e$, distinguishability $K_e$, and radius $t_{0,e}$ remain inside the aggregation. For $q=2$ and $nK_e\lambda_e\ge t_{0,e}^{-2}$, $\mathcal B_e(n)=c_e/(nK_e\lambda_e)$; outside that interior regime the radius truncates the bound.

\subsection{A Boundary-Region Classification Family}
\label{app:classification-mechanism}

Fix a known $0<\delta\le1/2$ and let $\mathcal P_\delta$ be the family in Equation~\eqref{eq:boundary-region-family}, with $X\sim\operatorname{Unif}[-1,1]$ and parameter rectangle $|\theta|,|\mu|\le\delta/8$. Define
\[
\mathfrak R_\delta(n)=\inf_{\widehat g_n}\sup_{|\theta|,|\mu|\le\delta/8}
\mathbb E_{\theta,\mu}[R_{\theta,\mu}(\widehat g_n)-R_{\theta,\mu}^\star],
\]
where the infimum ranges over measurable, possibly randomized learning rules based on $n$ independent observations.

\begin{appendixproposition}[Nuisance confounding near a classification boundary]
\label{prop:boundary-region-minimax}
At the reference parameter $(0,0)$, the raw and efficient task information satisfy
\[
J_{TT}=4\log3,
\qquad F_\delta=4\log\frac{2+\delta}{2-\delta},
\qquad 4\delta\le F_\delta\le\frac{64}{15}\delta.
\]
For every $n\ge1$,
\begin{equation}
\frac1{128e}\min\{\delta^2,(n\delta)^{-1}\}
\le\mathfrak R_\delta(n)
\le\min\left\{\frac{\delta^2}{32},
\frac{1-(1-\delta)^{n+1}}{2(n+1)\delta}\right\}.
\label{eq:boundary-region-finite-bounds}
\end{equation}
Consequently, the rate in Equation~\eqref{eq:boundary-region-minimax-rate} holds with constants independent of $n$ and $\delta$.
\end{appendixproposition}

\begin{proof}
Write $b_\delta(x)=\mathbf1\{|x|>\delta\}$ and $\eta_0(x)=1/2+x/4$. Throughout the rectangle, $1/8\le\eta_{\theta,\mu}\le7/8$, so the binary likelihood is regular in $(\theta,\mu)$ for each fixed $\delta>0$. Inside $|x|\le\delta$, the regression crosses $1/2$ at $-4\theta\in[-\delta/2,\delta/2]$. Outside, $|\theta+\mu|\le\delta/4<|x|/4$, so its sign relative to $1/2$ is the sign of $x$. Thus
\begin{equation}
g_{\theta,\mu}^\star(x)=\mathbf1\{x\ge-4\theta\},
\qquad
\mathcal E_{\theta,\mu}(g_a)=2(a-\theta)^2
\quad\text{for }g_a(x)=\mathbf1\{x\ge-4a\},\ |a|\le\delta/8.
\label{eq:boundary-region-risk-identity}
\end{equation}
The second identity integrates the margin over the interval between the two thresholds, which lies entirely inside $[-\delta,\delta]$.

\paragraph{Information.}
Let $v_0(x)=\eta_0(x)(1-\eta_0(x))=(4-x^2)/16$. The reference scores are
\[
s_T=\frac{Y-\eta_0(X)}{v_0(X)},\qquad s_N=b_\delta(X)s_T.
\]
Since $b_\delta^2=b_\delta$, $J_{TN}=J_{NN}>0$, so the nuisance-projection coefficient is one and $s_T^{\mathrm{eff}}=\mathbf1\{|X|\le\delta\}s_T$. Hence
\begin{equation}
J_{TT}=\int_{-1}^{1}\frac8{4-x^2}\,dx=4\log3,
\qquad
F_\delta=\int_{-\delta}^{\delta}\frac8{4-x^2}\,dx
=4\log\frac{2+\delta}{2-\delta}.
\label{eq:boundary-region-information-integrals}
\end{equation}
On $[-\delta,\delta]$, $2\le8/(4-x^2)\le32/15$, giving the asserted bounds on $F_\delta$.

\paragraph{Decision separation and distinguishability.}
For $0<t\le\delta/8$, take $P_{+t}=P_{t,-t}$ and $P_{-t}=P_{-t,t}$. Their regression functions are $\eta_0\pm t\mathbf1\{|x|\le\delta\}$ and their Bayes rules disagree on $(-4t,4t)$. The binary excess-risk identity therefore gives the exact separation
\begin{equation}
\inf_g\{\mathcal E_{+t}(g)+\mathcal E_{-t}(g)\}
=\frac12\int_{-4t}^{4t}\left(2t-\frac{|x|}{2}\right)dx
=4t^2.
\label{eq:boundary-region-separation}
\end{equation}
For Bernoulli parameters $p,q$, $\KL(\operatorname{Ber}(p)\Vert\operatorname{Ber}(q))\le(p-q)^2/[q(1-q)]$. Here the alternatives agree outside the central region; inside, $p-q=2t$ and $q\in[5/16,11/16]$. Since $P(|X|\le\delta)=\delta$ and $F_\delta\ge4\delta$,
\begin{equation}
\KL(P_{+t}\Vert P_{-t})
\le\frac{1024}{55}\delta t^2
\le\frac{256}{55}F_\delta t^2
<5F_\delta t^2.
\label{eq:boundary-region-kl}
\end{equation}
Theorem~\ref{thm:local-classification-difficulty} thus applies with $q=2$, $c_e=4$, $K_e=5$, $t_{0,e}=\delta/8$, and $\lambda_e=F_\delta$. Its proof gives
\[
\mathfrak R_\delta(n)\ge\frac1{2e}
\min\left\{\frac{\delta^2}{64},\frac1{5nF_\delta}\right\}
\ge\frac1{128e}\min\{\delta^2,(n\delta)^{-1}\},
\]
using $F_\delta\le64\delta/15$ in the last step.

\paragraph{Matching upper bound, including empty central samples.}
Let $I_i=\mathbf1\{|X_i|\le\delta\}$, $N=\sum_i I_i\sim\operatorname{Bin}(n,\delta)$, and define
\begin{equation}
\widehat\theta=\operatorname{clip}_{[-\delta/8,\delta/8]}
\left(\frac{\sum_{i=1}^nI_i(Y_i-1/2-X_i/4)}{N+1}\right).
\label{eq:boundary-region-estimator}
\end{equation}
This estimator is zero when $N=0$. Conditional on the feature observations, each selected residual has mean $\theta$, variance at most $1/4$, and independent noise. Before clipping, its mean-squared error is at most $(N/4+\theta^2)/(N+1)^2\le1/[4(N+1)]$; clipping cannot increase squared error because the interval contains $\theta$. The identity
\[
\mathbb E\frac1{N+1}
=\frac{1-(1-\delta)^{n+1}}{(n+1)\delta}
\]
and Equation~\eqref{eq:boundary-region-risk-identity} give the second upper bound in Equation~\eqref{eq:boundary-region-finite-bounds}. The fixed rule $g_0$ gives the first, since $2\theta^2\le\delta^2/32$. Taking the better of these two known rules proves the stated upper bound.
\end{proof}

Only the central observations distinguish the least-favorable task alternatives: their expected number is $n\delta$. The inverse-information rate is active once $n\delta^3$ is sufficiently large; otherwise the local decision scale $\delta^2$ caps the risk. The nuisance parameter here changes conditional probabilities outside the boundary region while preserving the Bayes rule at fixed $\theta$. This is a statistical learning mechanism, not a mathematical model of the image-background intervention.

\subsection{CVaR, Information Coverage, and Stabilization}
\label{app:theory-certificate}

For $0<\alpha<1$ and nonnegative $Z$, the CVaR minimization formula holds for integrable $Z$ \citep[Theorem~10]{rockafellar2002conditional}; the threshold-weight argument below gives its dual:
\begin{equation}
\begin{aligned}
\operatorname{CVaR}_\alpha(Z)
&=\inf_{\tau\in\mathbb R}\left\{\tau+\frac{\mathbb E[(Z-\tau)_+]}{1-\alpha}\right\}
=\sup_{\xi\in\mathcal W_\alpha}\mathbb E[\xi Z],\\
\mathcal W_\alpha&=\{\xi:0\le\xi\le(1-\alpha)^{-1},\ \mathbb E\xi=1\}.
\end{aligned}
\label{eq:app-cvar-dual}
\end{equation}
Both values are $+\infty$ if $\mathbb EZ=+\infty$. For integrable $Z$, a maximizing weight equals $(1-\alpha)^{-1}$ above $v_\alpha=\inf\{v:\Pr(Z\le v)\ge\alpha\}$, zero below it, and fills the remaining mass at the threshold. Such fractional weighting is possible because $\Pr(Z>v_\alpha)\le1-\alpha\le\Pr(Z\ge v_\alpha)$. The dual gives monotonicity, positive homogeneity, and $v_\alpha\le\operatorname{CVaR}_\alpha(Z)$. For the finite audit distribution $\Pi_G$, let $\pi_g=\Pi_G(\{g\})\ge0$, $\sum_g\pi_g=1$. Then
\begin{equation}
\operatorname{CVaR}_{\alpha,\pi}(Z_g)
=\max_{\substack{\sum_g\omega_g=1\\0\le\omega_g\le\pi_g/(1-\alpha)}}\sum_g\omega_gZ_g.
\label{eq:app-finite-cvar}
\end{equation}
These normalized tail weights, including threshold atoms, also govern Appendix~\ref{app:tail-transverse-influence}.

\begin{proof}[Proof of Proposition~\ref{prop:information-coverage}]
Let $\varepsilon>0$ and $\lambda_e\in[0,\infty)$ be finite almost surely. Write $L_e^{(\varepsilon)}=(\lambda_e+\varepsilon)^{-1}$ and $I_\alpha^{\FG}(\varepsilon)$ for the certificate in Equation~\eqref{eq:fisher-glass-certificate}. If $I_\alpha^{\FG}(\varepsilon)\ge\rho>\varepsilon$, then
\begin{equation}
\begin{aligned}
v_\alpha(L^{(\varepsilon)})&\le\operatorname{CVaR}_\alpha(L^{(\varepsilon)})
=[I_\alpha^{\FG}(\varepsilon)]^{-1}\le\rho^{-1},\\
\Pr_{e\sim\Pi}(\lambda_e\ge\rho-\varepsilon)
&=\Pr_{e\sim\Pi}(L_e^{(\varepsilon)}\le\rho^{-1})\ge\alpha.
\end{aligned}
\label{eq:app-information-coverage}
\end{equation}
The last step is the quantile definition and includes equality at the information threshold.
\end{proof}
Since $0<L^{(\varepsilon)}\le\varepsilon^{-1}$, always $I_\alpha^{\FG}(\varepsilon)\ge\varepsilon$; only $\rho>\varepsilon$ certifies a strictly positive floor. With four equally weighted groups and $\alpha=0.95$, one group already contains the entire upper-tail mass, so
\begin{equation}
\operatorname{CVaR}_{0.95}(L^{(\varepsilon)})=\max_gL_g^{(\varepsilon)},
\qquad I_{0.95}^{\FG}(\varepsilon)=\min_g\lambda_g+\varepsilon.
\label{eq:app-four-group-certificate}
\end{equation}

\paragraph{Exact limit and positivity.}
Set $L^{(0)}=\lambda^{-1}$, with $1/0=+\infty$. As $\varepsilon\downarrow0$, $L^{(\varepsilon)}\uparrow L^{(0)}$, and monotone convergence applies to $\mathbb E[\xi L^{(\varepsilon)}]$ for every feasible $\xi$. The supremum over $\xi$ commutes with the increasing supremum along any sequence $\varepsilon_k\downarrow0$, giving
\begin{equation}
\begin{aligned}
\operatorname{CVaR}_\alpha(L^{(\varepsilon)})&\uparrow\operatorname{CVaR}_\alpha(\lambda^{-1}),\\
I_\alpha^{\FG}(\varepsilon)&\downarrow I_\alpha^{\FG}(0)
:=\bigl[\operatorname{CVaR}_\alpha(\lambda^{-1})\bigr]^{-1},\qquad 1/(+\infty)=0.
\end{aligned}
\label{eq:app-exact-limit}
\end{equation}
For nonnegative $Z$, the feasible weight $\xi=1$ and variational choice $\tau=0$ imply
\begin{equation}
\mathbb EZ\le\operatorname{CVaR}_\alpha(Z)\le\frac{\mathbb EZ}{1-\alpha},
\qquad I_\alpha^{\FG}(0)>0\ \Longleftrightarrow\ \mathbb E[\lambda^{-1}]<\infty.
\label{eq:app-cvar-mean-bounds}
\end{equation}
Under the geometric tail assumptions, this is equivalent to $\beta=(c+\nu)/(2m)>1$; the exact certificate vanishes for $\beta\le1$. Information coverage is a population statement under $\Pi$, separate from the confidence level of an estimated lower certificate.

\section{Estimation and Certification}
\label{app:estimation-certification}
The bounds condition on a predictor, coordinates, calibration, and slices fixed independently of evaluation data. Evaluation-fitted normalization requires additional error control. Implementation details and numerical records appear in the artifact's \path{docs/extended_appendix_details/} (see the \hyperref[sec:artifact-availability]{artifact availability statement}).

\subsection{Operational Models and Matrix Targets}
\label{app:operational-matrices}
For evaluation distribution $P_g$ and calibrated $K$-class logits $\ell_w$, maps $\Phi_T(x)\in\mathbb R^{K\times d_T}$ and $\Phi_N(x)\in\mathbb R^{K\times d_N}$ convert task and nuisance parameters into logit changes:
\begin{equation}
\ell_{\theta,\mu}(x)=\ell_w(x)+\Phi_T(x)\theta+\Phi_N(x)\mu,
\qquad D=(\Phi_T\ \Phi_N),\quad d=d_T+d_N.
\label{eq:operational-perturbations}
\end{equation}
With $p=\operatorname{softmax}(\ell_w)$, $H=\diag(p)-pp^\top$, one-hot label vector $\mathbf e_Y$, and normalized feature coordinates $v=(v_T^\top\ v_N^\top)^\top$, distinguish
\begin{equation}
\begin{aligned}
M^{\mathrm{pred}}&=D^\top HD,\qquad M^{\mathrm{obs}}=ss^\top,
\quad s=D^\top(\mathbf e_Y-p),\\
M^{\mathrm{feat}}&=vv^\top,\qquad J_g=\E_{P_g}[M].
\end{aligned}
\label{eq:appB-matrix-targets}
\end{equation}
For a binary scalar logit $\ell_{\theta,\mu}=\ell_w+\theta z+\mu b$, this gives $M^{\mathrm{pred}}=p(1-p)vv^\top$, with $p=\sigma(\ell_w)$ and $v=(z,b)^\top$. Predictive scores have conditional mean zero under the fitted label distribution even when features do not; their Schur complement is efficient information for this operational model. Observed-label moments equal predictive Fisher under conditional correct specification, $P_g(Y=\cdot\mid X)=p(X)$; otherwise their centered covariance subtracts $\E[s]\E[s]^\top$. Feature geometry satisfies $J_g=\Sigma_g+\mu_g\mu_g^\top$, with $\mu_g=\E[v]$ and $\Sigma_g=\operatorname{Cov}(v)$. The prospective centered comparison uses within-slice covariance $\Sigma_g$ with divisor $n_g$ and unchanged global scales, removing the mean contribution without rescaling.

Prospective selection uses uncentered feature--Schur fractions; Appendix~\ref{app:prospective-intervention} compares centered and predictive targets. Within-group normalization of $p(1-p)$ weights changes the effective nuisance ridge and differs from the predictive moment. Hidden discovery leaves this factor unnormalized and shrinks toward the same-class pooled matrix, with $\eta=0.10$. Removing ridge leaves the declared matrix, including shrinkage, unchanged. Normalization conventions and nuisance-specification checks appear in Appendices~\ref{app:experimental-details} and~\ref{app:predefined-diagnostics}.

\paragraph{Operational subspaces and cost.}
A task coordinate specifies the predictive contrast to be audited; nuisance coordinates specify variations expected to preserve it. Candidate constructions include an answer-logit contrast with controlled prompt-template variation for a transformer, a region-level class contrast with validated appearance changes for segmentation, and a conditional denoising-score contrast for a diffusion model. Logits, probes, intermediate representations, and low-rank parameter perturbations provide finite-dimensional coordinates. For total operational dimension $d=d_T+d_N$, dense moment storage costs $O(d^2)$, and direct factorization or eigendecomposition costs $O(d^3)$ per slice. Minibatch accumulation and iterative eigensolvers support larger operational subspaces; feature or score extraction remains dependent on the backbone and input size. These examples are candidate extensions for future evaluation.

\paragraph{Admission of new nuisance specifications.}
A proposed admission protocol prespecifies independent checks of nuisance relevance, task preservation, split stability, and nuisance-block conditioning. An identifiability conclusion requires all four checks to pass; a failed prerequisite is recorded as an unvalidated specification. For a finite admitted family, report the minimum certificate with uncertainty accounting for the family. The present experiments evaluate prespecified nuisance hypotheses, with implemented checks and sensitivity analyses in Appendices~\ref{app:implemented-audit}, \ref{app:hidden-slice-discovery}, and~\ref{app:predefined-diagnostics}. The general admission procedure remains to be evaluated.

\subsection{Ridge Regularization and the Unregularized Target}
\label{app:ridge-unregularized}
For $J=\begin{psmallmatrix}A&B\\B^\top&C\end{psmallmatrix}\succeq0$, with $A=J_{TT}$, $B=J_{TN}$, $C=J_{NN}\succ0$, and $\lambda_{\rid}>0$, define
\begin{equation}
S_0(J)=A-BC^{-1}B^\top,\qquad
S_{\rid}(J)=A-B(C+\lambda_{\rid}I)^{-1}B^\top.
\label{eq:appB-schur-operators}
\end{equation}
Both are positive semidefinite. The resolvent identity, whose inverse factors commute, gives
\begin{equation}
\begin{aligned}
\Delta_{\rid}(J)&:=S_{\rid}(J)-S_0(J)
=\lambda_{\rid}BC^{-1}(C+\lambda_{\rid}I)^{-1}B^\top\succeq0,\\
\lVert\Delta_{\rid}(J)\rVert_{\op}
&\le\frac{\lambda_{\rid}M^2}{\kappa(\kappa+\lambda_{\rid})}
\quad\text{if }C\succeq\kappa I,\ \kappa>0,\ \lVert B\rVert_{\op}\le M.
\end{aligned}
\label{eq:ridge-exact-gap}
\end{equation}
Conditioning is essential: scalar blocks $A=1$, $B=\sqrt c$, $C=c>0$ give a positive-semidefinite joint matrix with $S_0=0$ and $S_{\rid}=\lambda_{\rid}/(c+\lambda_{\rid})\to1$ as $c/\lambda_{\rid}\downarrow0$. Comparisons therefore retain the declared standardized nuisance coordinates and ridge value.

\subsection{Error Allowances and Lower Certificates}
\label{app:lower-audit}
\begin{appendixproposition}[Simultaneous lower certificate]
\label{prop:appB-lower-certificate}
Fix $G$ groups and weights $\pi_g>0$ summing to one. Within each group, let $M_{gi}$ be i.i.d. with mean $J_g$, $0\preceq M_{gi}\preceq R_gI_d$, for specified finite $R_g$, and $\widehat J_g=n_g^{-1}\sum_iM_{gi}$. For $0<\delta<1$, set
\begin{equation}
\begin{aligned}
u_\delta&=\log(2dG/\delta),\qquad
 t_g=R_g\left(\sqrt{2u_\delta/n_g}+2u_\delta/(3n_g)\right),\\
\mathcal E_\delta&=\bigcap_g\{\lVert\widehat J_g-J_g\rVert_{\op}\le t_g\},
\qquad \Pr(\mathcal E_\delta)\ge1-\delta.
\end{aligned}
\label{eq:appB-matrix-event}
\end{equation}
Assume each $C_g\succ0$. Define observable bounds and allowances by
\begin{equation}
\begin{aligned}
\overline M_g&=\lVert\widehat B_g\rVert_{\op}+t_g,
&\underline\kappa_g&=\lambda_{\min}(\widehat C_g)-t_g,\\
h_{\rid}(t,M)&=t+(2Mt+t^2)/\lambda_{\rid}+M^2t/\lambda_{\rid}^2,
&a_g&=h_{\rid}(t_g,\overline M_g),\\
d_g&=\frac{\lambda_{\rid}\overline M_g^2}
{\underline\kappa_g(\underline\kappa_g+\lambda_{\rid})}
&&\text{when all }\underline\kappa_g>0.
\end{aligned}
\label{eq:appB-allowances}
\end{equation}
Report the lower certificate in~\eqref{eq:lower-certificate} with audit distribution $\pi$ only when these conditioning margins are positive; otherwise record conditioning failure. Its probability of exceeding the unregularized population certificate is at most $\delta$. On $\mathcal E_\delta$, a reported level $\underline I_{\alpha,\pi}\ge\rho>\varepsilon$ also implies $\sum_{g:\lambda_{\min}(S_0(J_g))\ge\rho-\varepsilon}\pi_g\ge\alpha$.
\end{appendixproposition}
\begin{proof}
For $X_{gi}=M_{gi}-J_g$, $\lVert X_{gi}\rVert_{\op}\le R_g$ and $\E[X_{gi}^2]=\E[M_{gi}^2]-J_g^2\preceq R_g^2I$. Matrix Bernstein~\citep[Theorem~1.4]{tropp2012user} with the displayed $t_g$ gives failure at most $\delta/G$ per group; the union bound allows dependence between group estimates. On this event, block extraction and Weyl give $\lVert B_g\rVert_{\op}\le\overline M_g$ and $C_g\succeq\underline\kappa_gI$, so~\eqref{eq:ridge-exact-gap} bounds the ridge gap by $d_g$.

For the estimation error, write $Q=(C+\lambda_{\rid}I)^{-1}$, $\widehat Q=(\widehat C+\lambda_{\rid}I)^{-1}$, and $\Delta B=\widehat B-B$. Positive semidefiniteness and the resolvent identity give $\lVert Q\rVert,\lVert\widehat Q\rVert\le\lambda_{\rid}^{-1}$ and $\lVert\widehat Q-Q\rVert\le t/\lambda_{\rid}^2$. Decompose the quadratic-block difference as $\Delta B\widehat Q\widehat B^\top+B\widehat Q\Delta B^\top+B(\widehat Q-Q)B^\top$. If $\lVert B\rVert\le M$, its three norms are bounded by $t(M+t)/\lambda_{\rid}$, $Mt/\lambda_{\rid}$, and $M^2t/\lambda_{\rid}^2$. Adding the task-block error $t$ proves the bound $h_{\rid}(t,M)$ and hence $a_g$.

Writing $\lambda_g^0=\lambda_{\min}(S_0(J_g))$, Weyl now yields
\begin{equation}
\lambda_g^0\ge\lambda_{\min}(S_{\rid}(J_g))-d_g
\ge\lambda_{\min}(S_{\rid}(\widehat J_g))-a_g-d_g.
\label{eq:appB-weyl}
\end{equation}
Since $S_0(J_g)\succeq0$, clipping in~\eqref{eq:lower-certificate} preserves $\underline\lambda_g\le\lambda_g^0$. Thus monotonicity of CVaR and reciprocation give
\begin{equation}
\underline I_{\alpha,\pi}\le
I^0_{\alpha,\pi}:=\left[\CVaR_{\alpha,g\sim\pi}
((\lambda_g^0+\varepsilon)^{-1})\right]^{-1}.
\label{eq:appB-certificate-guarantee}
\end{equation}
The information-coverage conclusion follows from Appendix~\ref{app:theory-certificate}.
\end{proof}
Confidence concerns held-out observations; coverage concerns the fixed weights $\pi$. A finite predeclared nuisance family can be included in the union bound over group--specification pairs; the infimum of certificates preserves coverage separately for each specification. Appendix~\ref{app:heldout-ratio} supplies matrix allowances for frozen soft gates, including random membership normalization, pooling, and support correction, to which the same deterministic argument applies.

\subsection{Implemented Audit and Numerical Comparison}
\label{app:implemented-audit}
Tables~\ref{tab:appB-conditioning}--\ref{tab:appB-training} audit uncentered feature moments of normalized scalar task and nuisance coordinates $z_i,b_i$, with uniform within-group weights. Separately from the finite-sample proposition, Gaussian/Wald simulation uses $10^5$ moment draws per group at \emph{nominal} 95\% confidence. For general normalized weights $w_i$ and $u_i=(z_i^2,z_ib_i,b_i^2)^\top$, draw $m^*=(A^*,B^*,C^*)^\top\sim N(\widehat m,\widehat\Sigma_{\widehat m})$, where
\begin{equation}
\widehat m=\sum_iw_iu_i,\qquad
\widehat\Sigma_{\widehat m}=
\frac{\sum_iw_i^2(u_i-\widehat m)(u_i-\widehat m)^\top}{1-\sum_iw_i^2}.
\label{eq:appB-wald-covariance}
\end{equation}
Uniform weights give sample covariance divided by $n$. Simulation fixes realized coordinate-standardization constants and weights, omitting evaluation-fitted standardization error. Bonferroni tail $\delta_{\mathrm W}=0.05/(3G)$ covers nuisance conditioning, ridge lower information, and an upper ridge gap within a run ($G=4$ Waterbirds; $18$ CivilComments).

Let $J^*$ have blocks $(A^*,B^*,C^*)$, and write $\widehat F_g^0=S_0(\widehat J_g)$, $\widehat F_g^{\rid}=S_{\rid}(\widehat J_g)$. The $\delta_{\mathrm W}$-quantiles $C_g^L,L_g^{\rid},L_g^0$ of $C^*,[S_{\rid}(J^*)]_+,[S_0(J^*)]_+$ are clipped to $[0,\widehat C_g]$, $[0,\widehat F_g^{\rid}]$, $[0,\widehat F_g^0]$, respectively. The $(1-\delta_{\mathrm W})$-quantile $U_g^\Delta$ of the closed-form ridge gap is enlarged to at least its plug-in value. Lower and upper quantiles use lower and higher order statistics, respectively. The implemented lower information is
\begin{equation}
\underline\lambda_g^{\mathrm W}
=\min\{[L_g^{\rid}-U_g^\Delta]_+,L_g^0,\widehat F_g^0\}.
\label{eq:appB-wald-lower}
\end{equation}
A denominator $C^*$ or $C^*+\lambda_{\rid}$ at or below $10^{-12}$ gives zero for its corresponding Schur value and an infinite gap; otherwise~\eqref{eq:ridge-exact-gap} is used even for non-PSD draws. An infinite upper gap or $C_g^L\le10^{-12}$ forces zero and an uncertifiable status; support, empirical-PSD, and identity checks can also prevent certification. With $\lambda_{\rid}=10^{-3}$, certificates with superscripts $0$, $\rid$, and $\mathrm W$ aggregate $\widehat F_g^0$, $\widehat F_g^{\rid}$, and $\underline\lambda_g^{\mathrm W}$, respectively, using uniform environment weights, $\alpha=0.95$, $\varepsilon=10^{-8}$, and fractional CVaR threshold-atom mass.

\begin{table}[H]
\centering\small
\caption{Same-cohort summaries of 75 default-ridge held-out audits. Conditioning/gap parentheses are within-run nominal Wald lower/upper values; entries are extrema across runs, not joint confidence statements across seeds or methods. Inflation is ridge minus exact plug-in certificate.}
\label{tab:appB-conditioning}
\label{tab:appB-certificates}
\setlength{\tabcolsep}{3pt}
\begin{tabular}{@{}p{.27\linewidth}rccccc@{}}
\toprule
& & \multicolumn{2}{c}{Conditioning and ridge gap} & \multicolumn{3}{c}{Certificates}\\
Scope & Audits & \shortstack{Minimum\\$J_{NN}$} & \shortstack{Maximum\\group gap} & \shortstack{Maximum\\inflation} & \shortstack{Minimum\\exact} & \shortstack{Minimum\\Wald lower}\\
\midrule
CivilComments diagnostics & 5 & \shortstack{1.456\\(1.417)} & \shortstack{0.00116\\(0.00133)} & 0.00018 & 0.4970 & 0.4618\\
Waterbirds diagnostics & 5 & \shortstack{0.211\\(0.178)} & \shortstack{0.00125\\(0.00166)} & 0.00071 & 0.1639 & 0.1504\\
Frozen-feature training & 30 & \shortstack{0.227\\(0.192)} & \shortstack{0.00144\\(0.00199)} & 0.00062 & 0.1972 & 0.1814\\
End-to-end training & 35 & \shortstack{0.349\\(0.349)} & \shortstack{0.00252\\(0.00300)} & 0.00140 & 0.2185 & 0.2004\\
\midrule
All audits & 75 & \shortstack{0.211\\(0.178)} & \shortstack{0.00252\\(0.00300)} & 0.00140 & 0.1639 & 0.1504\\
\bottomrule
\end{tabular}
\end{table}

Every run passed conditioning and satisfied $\underline I_{0.95}^{\mathrm W}\le\widehat I_{0.95}^{0}\le\widehat I_{0.95}^{\rid}$; Waterbirds diagnostic values are $0.1504$, $0.1639$, and $0.1646$. Across the five CivilComments diagnostic seeds, the mean ridge and unregularized certificates are $0.4993$ and $0.4991$, respectively. The inequality $\underline I_{0.95}^{\mathrm W}\le\widehat I_{0.95}^{0}$ is enforced by the caps and does not establish confidence coverage. Training audits use fresh models: original training runs lacked the retained environment blocks needed for checkpoint-specific auditing.

For the separate audits in Table~\ref{tab:appB-training}, frozen Group DRO + FG maximizes validation certificate within two percentage points of the best validation worst-group accuracy among $\{0.1,0.3\}$; seeds 17--21 select $(0.3,0.1,0.3,0.3,0.3)$. Repeated frozen-feature training/audit runs at ridges $10^{-4},10^{-3},10^{-2}$ give maximum group gaps $0.0001443$, $0.0014374$, $0.0138286$ and Wald upper values $0.0001996$, $0.0019880$, $0.0191165$.

\begin{table}[H]
\centering\small
\caption{Separate fresh training audits: five-seed means at the stated coefficients. Table~\ref{tab:appB-conditioning} instead includes frozen models selected over the full grid $\{0.01,0.03,0.1,0.3\}$.}
\label{tab:appB-training}
\begin{tabular}{@{}>{\raggedright\arraybackslash}p{.49\linewidth}ccc@{}}
\toprule
Training setting & Ridge plug-in & Exact plug-in & Wald lower\\
\midrule
Frozen FG, $\lambda_{\FG}=0.03$ & 0.4002 & 0.3999 & 0.3663\\
Frozen Group DRO + FG, selected from $\{0.1,0.3\}$ & 0.6254 & 0.6252 & 0.5664\\
End-to-end Group DRO + FG, $\lambda_{\FG}=10^{-4}$ & 0.5447 & 0.5442 & 0.4890\\
\bottomrule
\end{tabular}
\end{table}

\FloatBarrier

\section{Hidden-Slice Discovery}
\label{app:hidden-slice-discovery}
Appendix~\ref{app:discovery-results} reports this protocol's outcomes. Frozen-gate evaluation recomputes membership normalization and support correction. Prompts, model revision and execution details are in the \hyperref[sec:artifact-availability]{artifact}.

\subsection{Gate Family and Support Constraints}
\label{app:gate-support}
For the frozen standardized 576-dimensional representation $\psi(x)$ and task class $c$, define a logistic gate with parameters $\phi=(a_c,b_c)$:
\[
\begin{aligned}
h_{\phi,c}(x)&=\sigma(a_c^\top\psi(x)+b_c),\qquad
q_i^0=\frac{h_{\phi,c}(x_i)}{\sum_{j\in\mathcal D_c}h_{\phi,c}(x_j)},\quad
\sigma(t)=(1+e^{-t})^{-1},\\
m_{\mathrm{nom}}&=\tau_{\mathrm{slice}}n_c,\qquad
m_{\mathrm{ESS}}=\min\{n_c,\max\{1,0.8m_{\mathrm{nom}}\}\},\qquad
b_{\max}=\min\{1,2/m_{\mathrm{nom}}\}.
\end{aligned}
\]
Here $\mathcal D_c$ contains the current sample's examples with task label $c$, $n_c=|\mathcal D_c|$, and $\tau_{\mathrm{slice}}=0.10$ is within-class mass. Discovery calibrates $b_c$ to this mean membership; evaluation freezes the selected intercept, without re-centering or mass recalibration. The cap denominator is nominal slice size, not the ESS floor.

Support correction uses Euclidean projection of $q^0$ onto the capped simplex only if its maximum exceeds $b_{\max}$. Writing the result as $q^{\mathrm{cap}}$ and $u_i=1/n_c$, it takes $\widetilde q=u+t(q^{\mathrm{cap}}-u)$ with the largest $t\in[0,1]$ satisfying $\sum_i\widetilde q_i^2\le1/m_{\mathrm{ESS}}$. Thus $\widetilde q\ge0$, $\sum_i\widetilde q_i=1$, $\max_i\widetilde q_i\le b_{\max}$, and $n_{\eff}=(\sum_i\widetilde q_i^2)^{-1}\ge m_{\mathrm{ESS}}$. These constraints do not imply nuisance conditioning. Each sample uses its own $n_c$, making $\widetilde q$ batch-dependent. Reported mass averages raw selected-class membership; ESS uses $\widetilde q$.

\paragraph{Metrics and controls.}
\label{app:reported-slice-metrics}
Fisher and high-loss gates report accuracy, uncalibrated binary NLL (temperature one), and post-hoc composition averaged under $\widetilde q$, without information pooling. Predicted labels use a zero-logit threshold; calibrated NLL is stored separately. Random controls weight $k=\lceil\tau_{\mathrm{slice}}n_c\rceil$ sampled rows of the Fisher-selected class uniformly; the constant gate uses $1/n_c$. Controls using feature moments and centered covariances reevaluate the predictive-Fisher and high-loss gates without rediscovery.

\subsection{Discovery Objective and Settings}
\label{app:discovery-objective}
With matrix $M_i$ and same-class pool $\widehat J_{\mathrm{pool},c}=n_c^{-1}\sum_{i\in\mathcal D_c}M_i$, the ideal constrained objective searches over class--gate pairs:
\begin{equation}
\begin{aligned}
\widehat J_{\phi,c}^{(\eta)}&=(1-\eta)\sum_i\widetilde q_i(\phi)M_i+\eta\widehat J_{\mathrm{pool},c},\\
(c^\star,\phi^\star)&\in\argmax_{c,\phi}
\left\{\lambda_{\min}\!\left(S_{\rid}(\widehat J_{\phi,c}^{(\eta)})\right)+\varepsilon\right\}^{-1},
\qquad 0\le\eta<1.
\end{aligned}
\label{eq:hidden-slice-discovery}
\end{equation}
\begin{table}[H]
\centering\small
\caption{Fixed hidden-discovery settings; sensitivity grids are declared before metadata reveal.}
\label{tab:c-discovery-settings}
\begin{tabular}{@{}>{\raggedright\arraybackslash}p{.27\linewidth}>{\raggedright\arraybackslash}p{.68\linewidth}@{}}
\toprule
Component & Specification\\
\midrule
Representation/nuisance & Gate: frozen ImageNet MobileNetV3-small; nuisance: CLIP ViT-B/32 image projection onto a fixed water-minus-land text direction.\\
Head and calibration & Binary linear head: 1,200 full-batch AdamW steps, learning rate $0.01$, weight decay $10^{-4}$; temperature minimizes calibration NLL over 400 log-spaced values in $[0.05,20]$.\\
Gate optimization & Ten restarts; 1,000 AdamW steps, learning rate $0.03$, weight decay $10^{-4}$; Gaussian initial weights with standard deviation $0.02$.\\
Matrix and stabilization & $M_i=p_i(1-p_i)v_iv_i^\top$ on two task/nuisance coordinates; $\eta=0.10$, $\lambda_{\rid}=10^{-3}$, $\varepsilon=10^{-8}$.\\
Replication/sensitivity & Seeds 17--21; five validation folds; mass and shrinkage grids each $\{0.05,0.10,0.20\}$, headline $0.10$.\\
\bottomrule
\end{tabular}
\end{table}
For the scalar task/nuisance blocks, write $\widehat F^{\rid}=S_{\rid}(\widehat J_{\phi,c}^{(\eta)})$. Optimization and checkpoint/restart/class selection use $-\log(\widehat F^{\rid}+\varepsilon)-10(\bar h/\tau_{\mathrm{slice}}-1)^2$, where $\bar h=n_c^{-1}\sum_i h_{\phi,c}(x_i)$. The intercept is recalibrated before discovery evaluation; only mass-feasible finite checkpoints are retained, including a constant-gate baseline. High-loss search uses the same gate, mass and support rules, replacing negative log information with weighted, uncalibrated binary NLL.
Information uses $v_i=(z_i,b_i)^\top$: $z_i$ is the raw head logit $f_w(x_i)$ standardized by fitting-subset moments, $b_i$ the standardized CLIP nuisance score, and $p_i=\sigma(f_w(x_i)/T)$. Calibration changes the unnormalized predictive factor. Shrinkage pools complete matrices within the current sample's class before taking the Schur complement. Feature controls omit $p_i(1-p_i)$ but retain gate weights. The centered control forms gated and uniform same-class covariances separately, then pools them with weights $1-\eta$ and $\eta$ before taking the Schur complement. For CLIP, normalize five prompt embeddings per concept, normalize their concept means, then normalize the water-minus-land difference. Its dot product with normalized image embeddings gives the raw nuisance score, without background/group annotations.

\subsection{Fitting and Independent Evaluation}
\label{app:discovery-crossfit}
Table~\ref{tab:c-discovery-sequence} uses task labels and official splits throughout; background/group metadata remain sealed until all seeds finish held-out evaluation.
Jaccard overlap uses the $k=\max\{1,\lceil\tau_{\mathrm{slice}}n_c\rceil\}$ largest raw memberships in each selected class, intersected on original row identities; different classes give zero. None of the ten Fisher/high-loss gates has a cutoff tie. Fold stability applies this rule to the official training data's 15\% calibration subset: average same-class fold-pair overlaps, then weight these class averages by selected-fold counts normalized over classes with at least two gates. Neither comparison uses positive-membership supports or corrected-weight ranks.

\begin{table}[htbp]
\centering\small
\caption{Fitting and evaluation sequence under the frozen protocol.}
\label{tab:c-discovery-sequence}
\setlength{\tabcolsep}{4pt}
\begin{tabular}{@{}>{\raggedright\arraybackslash}p{.18\linewidth}>{\raggedright\arraybackslash}p{.50\linewidth}>{\raggedright\arraybackslash}p{.27\linewidth}@{}}
\toprule
Stage & Fitted or computed & Carried forward\\
\midrule
Training/calibration & Split official training data 85/15 by task label: fit head and coordinate normalizers on 85\%; temperature on 15\%. CLIP scores use official-training, then fit-subset normalization. & Fixed predictor, coordinates, nuisance direction, temperature.\\
Validation folds & Five shared task-stratified assignments across selectors; discover on complementary folds, evaluate on held-out folds; assess stability on the 15\% training-calibration subset. & Protocol assessment; fold gates are not ensembled.\\
Full validation & Refit the gate under the fixed protocol; select class, gate, and restart by discovery utility. & Freeze gate weights, calibrated intercept, selected class, support settings.\\
Test & Apply frozen memberships; recompute admissible weights, same-class pool, and metrics before metadata reveal. & Held-out results, then post-hoc interpretation.\\
\bottomrule
\end{tabular}
\par\smallskip
\begin{minipage}{\linewidth}\normalsize
Support correction leaves all five primary Fisher gates' test weights unchanged; for high-loss seed 19 it raises ESS from $94.48$ to $102.72$. Frozen memberships need not give fixed numerical weights or inactive bootstrap correction.
\end{minipage}
\end{table}

\subsection{Held-Out Targets and Uncertainty}
\label{app:heldout-ratio}
Conditional on fitting data and fitted quantities, let $P_c$ be the selected-class distribution and $n=n_c$ its evaluation sample size. For frozen gate $h$ and matrix map $M$,
\begin{equation}
J_h=\frac{A}{p},\quad A=\E_{P_c}[hM],\quad p=\E_{P_c}[h]>0;
\qquad \widehat J_h=\frac{\widehat A}{\widehat p},\quad
(\widehat A,\widehat p)=\frac1n\sum_{i=1}^n(h_iM_i,h_i).
\label{eq:hidden-slice-population-target}
\end{equation}
\begin{appendixproposition}[Frozen-gate ratio bound]
\label{prop:frozen-gate-ratio}
For i.i.d. evaluation observations from $P_c$, assume $0\le h\le1$ and $0\preceq M\preceq M_{\max}I_d$, with finite $M_{\max}$ specified independently of evaluation data. Given $0<\delta<1$, let
\[
u=\log(4d/\delta),\qquad
r_A=M_{\max}\left(\sqrt{2u/n}+2u/(3n)\right),\qquad
r_p=\sqrt{\log(4/\delta)/(2n)}.
\]
With conditional probability at least $1-\delta$, simultaneously $\lVert\widehat A-A\rVert_{\op}\le r_A$ and $|\widehat p-p|\le r_p$. Whenever $\underline p=\widehat p-r_p>0$ on this event,
\begin{equation}
\lVert\widehat J_h-J_h\rVert_{\op}
\le r_J:=\frac{r_A+\lVert\widehat J_h\rVert_{\op}r_p}{\widehat p-r_p}.
\label{eq:frozen-gate-ratio-radius}
\end{equation}
\end{appendixproposition}
\begin{proof}
The centered matrices $h_iM_i-A$ have norm at most $M_{\max}$ and variance bounded by $M_{\max}^2I_d$. Matrix Bernstein~\citep[Theorem~1.4]{tropp2012user} gives the numerator event with failure at most $\delta/2$; Hoeffding gives the denominator event with failure at most $\delta/2$. Their union bound and
\[
\widehat J_h-J_h=
\frac{(\widehat A-A)-\widehat J_h(\widehat p-p)}{p},
\qquad p\ge\widehat p-r_p>0,
\]
prove the observable bound. The shared random denominator is therefore explicitly controlled.
\end{proof}

\paragraph{Pooling and support correction.}
Set $J_{\mathrm{pool},c}=\E_{P_c}[M]$ and $J_h^{(\eta)}=(1-\eta)J_h+\eta J_{\mathrm{pool},c}$. Allocate $\delta_{\mathrm{ratio}}+\delta_{\mathrm{pool}}=\delta$, using the former in Proposition~\ref{prop:frozen-gate-ratio}. For the evaluation mean $\widehat J_{\mathrm{pool},c}=n^{-1}\sum_iM_i$, matrix Bernstein supplies
\[
r_{\mathrm{pool}}=M_{\max}\left[\sqrt{\frac{2\log(2d/\delta_{\mathrm{pool}})}n}
+\frac{2\log(2d/\delta_{\mathrm{pool}})}{3n}\right].
\]
Writing $\widetilde J_h=\sum_i\widetilde q_iM_i$, the triangle inequality gives
$\lVert\widetilde J_h-\widehat J_h\rVert_{\op}\le M_{\max}\sum_i|\widetilde q_i-q_i^0|=:r_{\mathrm{proj}}$.
Hence, on the simultaneous event and when $\underline p>0$,
\begin{equation}
\lVert(1-\eta)\widetilde J_h+\eta\widehat J_{\mathrm{pool},c}-J_h^{(\eta)}\rVert_{\op}
\le(1-\eta)(r_J+r_{\mathrm{proj}})+\eta r_{\mathrm{pool}}.
\label{eq:gate-projection-allowance}
\end{equation}
The union bound permits dependence between pool and ratio estimates; $r_{\mathrm{proj}}$ covers batch-dependent support correction. If the pool is fixed from fitting data, it belongs to the conditional target and needs no evaluation-error term. Appendix~\ref{app:lower-audit} converts the matrix allowance into a Schur/eigenvalue lower certificate, requiring positive lower membership mass and nuisance-conditioning margins. The target retains shrinkage; evaluation-estimated centering additionally requires first-moment control.

\paragraph{Reported uncertainty.}
Each of 2,000 within-class bootstrap replicates resamples examples with membership, coordinates, logit and label together, recomputing support-corrected weights and the same-class pool. Gate coefficients/intercept, normalizers and temperature stay frozen. The 95\% interval uses the 2.5th/97.5th percentiles of the pooled predictive matrix's ridge Schur value, retaining $\eta=0.10$ and $\lambda_{\rid}=10^{-3}$, with linear quantile interpolation; $\varepsilon$ enters inverse information separately. For any reported metric $T$, form paired differences $\Delta_s=T_s^{\mathrm{gate}}-T_s^{\mathrm{control}}$ across five seeds, then report $\overline\Delta\pm t_{0.975,4}s_\Delta/\sqrt5$, with sample standard deviation $s_\Delta$; single-procedure intervals use its five values. Bootstrap intervals condition on fitted quantities; seed intervals reflect shared-benchmark fitting variation. Neither validates the finite-sample radii above.

\FloatBarrier

\clearpage
\section{Experimental Protocols and Additional Results}
\label{app:experimental-details}
\subsection{Experimental Quantities and Aggregation}
\label{app:experimental-conventions}

Table~\ref{tab:d-experiment-overview} identifies the operational targets; Appendix~\ref{app:operational-matrices} defines the matrix constructions. Real-data audits use nuisance ridge $10^{-3}$ and stabilization $10^{-8}$; synthetic audits use $\varepsilon=10^{-12}$. Certificates use $\alpha=0.95$. Primary predefined-group coordinates use training moments; prospective normalization is specified below. Reported evaluation Loss-CVaR averages group-mean NLLs at or above their empirical 95th percentile, weighting groups equally; synthetic experiments use environment-mean NLLs. The Loss-CVaR training baseline averages the largest $\lceil0.1n\rceil$ individual-example NLLs.

\begin{table}[htbp]
\centering\small
\setlength{\tabcolsep}{3pt}
\caption{Diagnostic targets and replication units.}
\label{tab:d-experiment-overview}
\begin{tabular}{@{}p{.26\linewidth}p{.61\linewidth}c@{}}
\toprule
Experiment & Representation and reported quantity & Seeds \\
\midrule
Prospective Waterbirds & Fine-tuned MobileNetV3-small; background-probe surviving fraction & 22--26 \\
Synthetic & Scalar task $z=s+\delta_e r$ and nuisance $s$; exact population moments, inverse-information tail and certificate & 11--15 \\
Hidden Waterbirds slices & Frozen MobileNetV3-small and fixed CLIP direction; shrunken predictive information & 17--21 \\
CivilComments groups & TF-IDF logistic classifier; feature--Schur certificate & 13--17 \\
Waterbirds groups & Frozen MobileNetV3-small and logistic fitting; feature--Schur certificate & 17--21 \\
\bottomrule
\end{tabular}
\par\smallskip
\end{table}

\paragraph{Uncertainty.}
Prospective, synthetic, predefined-group, and frozen-feature training summaries report nominal 95\% normal-approximation half-widths $1.96s/\sqrt5$, with sample standard deviation $s$. Hidden-slice and group-free comparisons use five-seed Student-$t$ intervals; contrasts are paired within seed. The end-to-end coefficient sweep reports standard deviations, and the conditional Gaussian/Wald audit is separate (Appendix~\ref{app:implemented-audit}). Hidden-gate bootstrap intervals condition on the frozen gate and recompute evaluation weights (Appendix~\ref{app:hidden-slice-discovery}). Acquisition retains means without recovered per-seed intervals.

Seeds describe fitting/partition or control variation on fixed benchmark splits, not independent datasets. For the predefined-group Waterbirds diagnostic in Appendix~\ref{app:predefined-diagnostics}, the classifier is deterministic after feature extraction and logistic fitting; only its random controls and reserve anchor vary across seeds. Prospective intervals use training seeds: 23,176 composites share 5,794 birds, with two donors per bird and arm. Accuracy changes are percentage points; information retains its original units.

\subsection{Prospective Nuisance Intervention}
\label{app:prospective-intervention}

\paragraph{Training and frozen selection.}
ImageNet-initialized MobileNetV3-small~\citep{russakovsky2015imagenet,howard2019searching} is fully fine-tuned on Waterbirds~\citep{sagawa2019distributionally} for seeds 22--26: 50 epochs, batch size 32, learning rate $10^{-4}$, and bf16 on an H100. A validation-fitted temperature and class-balanced logistic background probe use the pooled representation. Five probe folds are stratified by validation group (3,000 maximum iterations); features use validation standardization. Validation nuisance scores are out of fold and use validation moments; test scores use the full-validation probe and test moments. Task logits use training moments. Test normalization never selects the slice. The rule is
\begin{equation}
\begin{aligned}
\mathcal G_{\mathrm{eligible}}=\{g:\;&\operatorname{Acc}_{v}(g)\ge\operatorname{Acc}_{v},\quad
\operatorname{NLL}_{v}(g)\le\operatorname{median}_{h}\operatorname{NLL}_{v}(h)\},\\
g^\star&\in\underset{g\in\mathcal G_{\mathrm{eligible}}}{\arg\min}
\lambda_{\min}(\widehat F_g^{\rid})/\widehat J_{TT,g}.
\end{aligned}
\label{eq:d-prospective-selection}
\end{equation}
Here $v$ denotes validation and eligibility uses uncalibrated NLL. Temperature scaling~\citep{guo2017calibration} minimizes full-validation binary NLL over 400 log-spaced temperatures in $[0.05,20]$, choosing the smaller temperature at ties. The checkpoint, temperature, probe, slice and audit constants are frozen before scoring; checkpoint mismatches abort evaluation. Table~\ref{tab:d-validation-performance} shows that LL (landbird/land) and WW (waterbird/water) qualify; WW is selected in all five seeds.

\begin{table}[htbp]
\centering\small
\setlength{\tabcolsep}{3pt}
\caption{Validation selection. Means $\pm$ nominal 95\% normal-approximation half-widths across seeds 22--26; accuracy is a fraction. Information uses uncentered feature moments.}
\label{tab:d-validation-performance}
\label{tab:d-validation-information}
\textit{(a) Performance and eligibility}\par\smallskip
\begin{tabular}{@{}lrccc@{}}
\toprule
Group & $n$ & Accuracy & NLL & Eligible \\
\midrule
Landbird/land & 467 & $0.990\pm0.005$ & $0.038\pm0.016$ & Yes \\
Landbird/water & 466 & $0.703\pm0.019$ & $1.670\pm0.156$ & No \\
Waterbird/land & 133 & $0.390\pm0.021$ & $7.718\pm0.800$ & No \\
Waterbird/water & 133 & $0.920\pm0.013$ & $0.638\pm0.038$ & Yes \\
\bottomrule
\end{tabular}
\par\smallskip
\textit{(b) Information}\par\smallskip
\begin{tabular}{@{}lccc@{}}
\toprule
Group & Raw information & Projected information & Surviving fraction \\
\midrule
Landbird/land & $0.737\pm0.046$ & $0.436\pm0.045$ & $0.590\pm0.027$ \\
Landbird/water & $0.794\pm0.032$ & $0.289\pm0.014$ & $0.365\pm0.026$ \\
Waterbird/land & $0.973\pm0.033$ & $0.764\pm0.031$ & $0.786\pm0.036$ \\
Waterbird/water & $2.063\pm0.054$ & $0.511\pm0.021$ & $0.248\pm0.015$ \\
\bottomrule
\end{tabular}
\par\smallskip
\end{table}

\paragraph{Foreground-preserving construction.}
Waterbirds combines CUB birds~\citep{wah2011cub} with Places backgrounds~\citep{zhou2018places} under soft CUB mattes. For each bird, we retain background pixels verbatim outside the dilated foreground and fill the covered region using Telea inpainting~\mbox{\citep{telea2004image}}, with a 5-pixel inpainting radius and a $5\times5$ elliptical dilation kernel. Its extracted bird layer is recomposed onto training-image donors of the same or opposite land/water background type. Donors have foreground fraction at most $0.08$; a deterministic sequence tries four, taking the first with visible inpainted area at most $0.03$, or the least-inpainted of four. The same-type arm undergoes identical recomposition.

Both arms use the same foreground layer, matte, label and source birds. Opaque foreground pixels are unchanged before JPEG encoding (quality 95, no chroma subsampling); soft-edge mixtures vary with background. Two draws per arm yield 23,176 composites. Visible inpainted area has mean $4.13\%$, p95 $8.45\%$, maximum $20.37\%$, and arm means $4.13\%$/$4.12\%$. Correctness is averaged across donors within bird, then across birds.

\paragraph{Outcomes and confident flips.}
Table~\ref{tab:d-intervention-outcomes} reports all groups. The confident half uses the within-slice median calibrated true-label probability among originally correct birds, including ties. In WW, opposite-type/control flip rates are $56.4\%$/$3.3\%$ overall and $41.8\%$/$0.7\%$ in that half; the fixed threshold $p\ge0.9$ selects no WW birds.

\begin{table}[htbp]
\centering\small
\setlength{\tabcolsep}{3pt}
\caption{Held-out interventions. Accuracies and flip rates are seed means in percent; controlled drops are same-type minus opposite-type percentage points with nominal 95\% normal-approximation half-widths. Panel (b) counts are mean confident-half counts.}
\label{tab:d-intervention-outcomes}
\label{tab:d-confident-flips}
\textit{(a) Paired intervention arms}\par\smallskip
\begin{tabular}{@{}lrcccc@{}}
\toprule
Group & Birds & Original & Same type & Opposite type & Controlled drop \\
\midrule
Landbird/land & 2,255 & 98.9 & 99.3 & 72.6 & $26.7\pm1.8$ \\
Landbird/water & 2,255 & 71.3 & 72.3 & 99.2 & $-26.9\pm2.0$ \\
Waterbird/land & 642 & 42.3 & 40.6 & 92.3 & $-51.7\pm2.0$ \\
\textbf{Waterbird/water} & 642 & \textbf{92.0} & \textbf{93.2} & \textbf{40.6} & $\mathbf{52.6\pm1.4}$ \\
\bottomrule
\end{tabular}
\par\smallskip
\textit{(b) Opposite-type flips among originally correct predictions}\par\smallskip
\begin{tabular}{@{}lccc@{}}
\toprule
Group & \shortstack{All originally\\correct} & \shortstack{Confident\\half} & \shortstack{Confident-half\\count} \\
\midrule
Landbird/land & 26.8 & 14.8 & 1,122 \\
Landbird/water & 0.2 & 0.1 & 805 \\
Waterbird/land & 1.2 & 0.3 & 136 \\
Waterbird/water & 56.4 & 41.8 & 299 \\
\bottomrule
\end{tabular}
\par\smallskip
\end{table}

\paragraph{Matched selection comparisons.}
Reapplying eligibility within each seed leaves LL and WW. Minimum validation accuracy and maximum uncalibrated NLL, evaluated separately on this identical set, agree with the surviving-fraction selector in all five seeds. All-group comparisons answer a different question; landbird/water determines the model-level certificate (Table~\ref{tab:d-selection-comparison}).
\paragraph{Centering and predictive Fisher.}
The post hoc check fixes the checkpoints, temperatures, validation coordinates $v=(z,b)^\top$, original scales, eligible sets and intervention outcomes. No model/probe is refitted. With $\langle\cdot\rangle_g$ the divisor-$n_g$ slice mean, compare
\[
J_g=\langle vv^\top\rangle_g=\Sigma_g+\mu_g\mu_g^\top,\quad
\Sigma_g=\langle(v-\mu_g)(v-\mu_g)^\top\rangle_g,\quad \mu_g=\langle v\rangle_g.
\]
Centering does not rescale. The local model $\operatorname{logit}p_{\theta,\xi}(x)=f_w(x)/\widehat T+\theta z(x)+\xi b(x)$ at zero has score $(Y-p)v$ with conditional mean zero even when $\mu_g\ne0$. Its predictive Fisher and the common fraction are
\[
I_g=\langle p(1-p)vv^\top\rangle_g,\qquad
r_\rho(A)=1-\frac{A_{zb}^2}{A_{zz}(A_{bb}+\rho)},\quad\rho=10^{-3}.
\]
Perturbations follow calibration and profile only the declared background coordinate. The normalized-weight comparator is $K_g=I_g/\bar q_g$, $\bar q_g=\langle p(1-p)\rangle_g$, so $r_\rho(K_g)=r_{\bar q_g\rho}(I_g)$: normalization preserves the unregularized fraction but changes effective ridge. Table~\ref{tab:d-matrix-sensitivity} therefore reports it separately.
\begin{table}[htbp]
\centering\small
\setlength{\tabcolsep}{3pt}
\caption{Selection and matrix sensitivity, evaluated within each seed. Panel (a) distinguishes original eligible and all-group sets; drops are mean percentage points. Panel (b) fixes the original eligible LL/WW set and interventions; fractions/drops are means $\pm$ nominal 95\% normal-approximation half-widths.}
\label{tab:d-selection-comparison}
\label{tab:d-matrix-sensitivity}
\textit{(a) Matched selectors}\par\smallskip
\begin{tabular}{@{}p{.39\linewidth}p{.13\linewidth}p{.30\linewidth}r@{}}
\toprule
Statistic & Candidates & Selected group, seeds & Drop \\
\midrule
Minimum raw task information & Eligible & Landbird/land, 5/5 & 26.7 \\
Minimum absolute projected information & Eligible & Landbird/land, 5/5 & 26.7 \\
Minimum surviving fraction & Eligible & Waterbird/water, 5/5 & 52.6 \\
Minimum validation accuracy or maximum validation NLL & Eligible & Waterbird/water, 5/5 & 52.6 \\
Minimum normalized-weight fraction & Eligible & Waterbird/water, 5/5 & 52.6 \\
Maximum NLL or minimum accuracy & All four & Waterbird/land, 5/5 & $-51.7$ \\
Group determining $I_{0.95}^{\FG}$ & All four & Landbird/water, 5/5 & $-26.9$ \\
\bottomrule
\end{tabular}
\par\smallskip
\textit{(b) Frozen-coordinate constructions}\par\smallskip
\begin{tabular}{@{}lcccc@{}}
\toprule
Construction & LL fraction & WW fraction & Selected, seeds & Drop \\
\midrule
Uncentered $J_g$ & $0.590\pm0.027$ & $0.248\pm0.015$ & WW, 5/5 & $52.6\pm1.4$ \\
Centered $\Sigma_g$ & $0.779\pm0.028$ & $0.855\pm0.032$ & LL, 5/5 & $26.7\pm1.8$ \\
Predictive Fisher $I_g$ & $0.987\pm0.014$ & $0.273\pm0.016$ & WW, 5/5 & $52.6\pm1.4$ \\
Normalized weights $K_g$ & $0.987\pm0.014$ & $0.269\pm0.016$ & WW, 5/5 & $52.6\pm1.4$ \\
\bottomrule
\end{tabular}
\par\smallskip
\end{table}

Centering selects LL in all seeds, whereas predictive Fisher retains WW. Slice means therefore contribute to the original ranking: for WW in seed 22, $\mu_z^2=2.094$ versus $\Sigma_{zz}=0.0636$. Table~\ref{tab:d-matrix-per-seed} preserves the per-seed reversal; full means, covariance entries, weights, matrices and hashes are retained in the artifact.

\begin{table}[htbp]
\centering\small
\caption{Per-seed fractions on the original global scales. Each cell is LL / WW; the smaller value determines selection. Centering reverses the ranking in every seed.}
\label{tab:d-matrix-per-seed}
\begin{tabular}{@{}rccc@{}}
\toprule
Seed & Uncentered $r_\rho(J_g)$ & Centered $r_\rho(\Sigma_g)$ & Predictive $r_\rho(I_g)$ \\
\midrule
22 & 0.5570 / 0.2291 & 0.7430 / 0.8542 & 0.9817 / 0.2587 \\
23 & 0.6048 / 0.2494 & 0.8098 / 0.8481 & 0.9920 / 0.2729 \\
24 & 0.6356 / 0.2755 & 0.8132 / 0.8359 & 0.9620 / 0.3032 \\
25 & 0.5803 / 0.2392 & 0.7514 / 0.9156 & 0.9998 / 0.2571 \\
26 & 0.5716 / 0.2460 & 0.7779 / 0.8189 & 0.9998 / 0.2732 \\
\bottomrule
\end{tabular}
\end{table}

\paragraph{Development chronology.}
The surviving-fraction rule followed an absolute-information pilot on seeds 17--21 and was frozen for newly trained seeds 22--26 on the same benchmark. Primary-cohort entries use the documented rerun; the original-versus-rerun ledger is retained in the artifact. Appendix~\ref{app:extended-training} gives the Fisher-Glass training comparison under this same intervention.

\paragraph{Raw, projected, and tail summaries.}
We compare four summaries of retained validation geometry: mean raw task second moment $A$, arithmetic mean projected information $P$, reciprocal mean inverse information $H$, and inverse-CVaR certificate $C$. For the four equally weighted groups, let $F_g=\widehat F_g^{\rid}$ and $L_g=(F_g+\varepsilon)^{-1}$. Then
\[
\begin{aligned}
A&=\tfrac14\sum_g\widehat J_{TT,g}, &
P&=\tfrac14\sum_g F_g,\\
H&=\left(\tfrac14\sum_g L_g\right)^{-1}, &
C&=\left[\CVaR_{0.95,\pi}(L_g)\right]^{-1},\qquad \pi_g=\tfrac14.
\end{aligned}
\]
Thus $H$ is the harmonic mean of the stabilized projected information values $F_g+\varepsilon$. The four summaries use uncentered feature moments; projection uses nuisance ridge $10^{-3}$, and $H$ and $C$ use reciprocal stabilization $\varepsilon=10^{-8}$. This post hoc analysis joins validation summaries to held-out interventions in the primary (seeds 22--26) and development (17--21) cohorts. It compares summaries of existing models, not training objectives. ``Selected FG'' denotes the Fisher-Glass models selected within each cohort, including the accuracy-budget fallbacks listed in the caption; Appendix~\ref{app:extended-training} details the primary training comparison.

\begin{table}[htbp]
\centering\small
\setlength{\tabcolsep}{3pt}
\caption{Four validation summaries and their intervention associations. Panels (a)--(b): means $\pm$ nominal 95\% normal-approximation half-widths, $1.96s/\sqrt5$; changes are paired within seed and drops are signed percentage points. Selected FG includes accuracy-budget fallbacks (primary seeds 22, 24; development seed 20). Panel (c): descriptive Spearman coefficients for ten models arranged in five ERM/FG seed pairs per cohort, with average ranks for tied outcomes.}
\label{tab:four-way-swap}
\textit{(a) Model summaries and paired changes}\par\smallskip
\begin{tabular}{@{}llcccc@{}}
\toprule
Cohort & Model/change & $A$: raw & $P$: projected & $H$: harmonic & $C$: inverse-CVaR\\
\midrule
Primary & ERM & $1.142\pm0.012$ & $0.500\pm0.010$ & $0.442\pm0.011$ & $0.289\pm0.014$\\
& Selected FG & $0.730\pm0.086$ & $0.484\pm0.047$ & $0.435\pm0.043$ & $0.330\pm0.043$\\
& FG $-$ ERM & $-0.412\pm0.079$ & $-0.016\pm0.046$ & $-0.007\pm0.041$ & $+0.041\pm0.046$\\
Development & ERM & $1.152\pm0.005$ & $0.489\pm0.011$ & $0.413\pm0.014$ & $0.255\pm0.018$\\
& Selected FG & $0.754\pm0.118$ & $0.519\pm0.065$ & $0.433\pm0.057$ & $0.303\pm0.082$\\
& FG $-$ ERM & $-0.398\pm0.118$ & $+0.030\pm0.063$ & $+0.020\pm0.066$ & $+0.048\pm0.095$\\
\bottomrule
\end{tabular}
\par\smallskip
\textit{(b) ERM group-selection rules and controlled drops}\par\smallskip
\begin{tabular}{@{}>{\raggedright\arraybackslash}p{.34\linewidth}llrr@{}}
\toprule
Rule & Candidates & Selected, each cohort & Primary & Development\\
\midrule
Minimum raw second moment & All four & LL 4/5; LW 1/5 & $+15.8\pm21.3$ & $+16.9\pm23.2$\\
Minimum projected / largest inverse contribution / CVaR-active group & All four & LW 5/5 & $-26.9\pm2.0$ & $-29.7\pm2.2$\\
Largest inverse contribution & Eligible & LL 5/5 & $+26.7\pm1.8$ & $+29.2\pm2.8$\\
\bottomrule
\end{tabular}
\par\smallskip
\textit{(c) Model-level rank associations with held-out outcomes}\par\smallskip
\begin{tabular}{@{}lrrrr@{}}
\toprule
& \multicolumn{2}{c}{Primary} & \multicolumn{2}{c}{Development}\\
Summary & $D_{\max}$ & $W_{\mathrm{opp}}$ & $D_{\max}$ & $W_{\mathrm{opp}}$\\
\midrule
$A$: mean raw & $-0.10$ & $+0.26$ & $+0.26$ & $-0.13$\\
$P$: mean projected & $-0.09$ & $+0.21$ & $-0.03$ & $-0.43$\\
$H$: reciprocal mean inverse & $-0.15$ & $+0.21$ & $-0.12$ & $-0.40$\\
$C$: inverse-CVaR & $+0.15$ & $-0.24$ & $-0.03$ & $-0.48$\\
\bottomrule
\end{tabular}
\end{table}

For Table~\ref{tab:four-way-swap}(c), $D_{\max}=100\max_g(\operatorname{Acc}_{\mathrm{same},g}-\operatorname{Acc}_{\mathrm{opposite},g})$ and $W_{\mathrm{opp}}=\min_g\operatorname{Acc}_{\mathrm{opposite},g}$, both over all four groups. Integer correct counts recover exact outcome ties: primary ERM seeds 22 and 25 share drop $348/642$, despite different floating-point subtractions. LL and LW denote landbird/land and landbird/water. Negative drops mean opposite-type recomposition improves accuracy. The group with the largest inverse-information contribution accounts for $38.3\%$ and $40.6\%$ of summed inverse information on average in the primary and development cohorts, respectively. The original headline rules differed: surviving fraction in the primary cohort, absolute projected information in development; this comparison consistently uses all four groups for model summaries and correlations.

Projection and aggregation give different model summaries, with rank associations varying across cohorts. For four equally weighted groups and $\alpha=0.95$, $C=\min_gF_g+\varepsilon$; its active group also minimizes projected information and maximizes inverse contribution. Here that group is LW, whose opposite-type recomposition improves accuracy. The prospective WW diagnosis instead uses the surviving fraction within the eligible set. This comparison does not establish superior swap-damage prediction by inverse-CVaR; paired changes in $P,H,C$ all have intervals containing zero. The artifact provides per-model and per-seed values, exact-count outcomes, input hashes, and replay code.

\paragraph{Auxiliary flip prediction.}
Each originally correct source bird contributes two opposite-type donor rows. All predictor covariates are evaluated on the original, pre-intervention image. Baseline predictors are uncalibrated NLL, calibrated true-label confidence, absolute raw logit and standardized task logit $z$. Augmentation adds $\beta_gb$ and $|z-\beta_gb|$, with $\beta_g=\widehat J_{TN,g}/(\widehat J_{NN,g}+10^{-3})$ frozen from the validation block. Both models use identical five-fold, nonshuffled CUB-species \texttt{GroupKFold} splits, keeping species and donor pairs together. Logistic regression uses an intercept, $\ell_2$ penalty, $C=1$, L-BFGS, tolerance $10^{-4}$, maximum 5,000 iterations, seed 0 and no class weighting or tuning. All columns, including group indicators, are centered and scaled using training-fold means and population standard deviations; standard deviations below $10^{-9}$ are replaced by one.

Pool out-of-fold predictions within seed before computing average precision (reported as AUPRC), area under the receiver operating characteristic curve (AUROC), and mean log likelihood. For seed $s$, $N_s$ counts all retained donor rows in the analyzed population, including flips and nonflips. Let $y^{\mathrm{flip}}_{si}=1$ indicate an incorrect prediction on the donor composite and $y^{\mathrm{flip}}_{si}=0$ a correct prediction. The probabilities $q^A_{si}$ and $q^B_{si}$ are the augmented and baseline out-of-fold flip predictions. For clipped probabilities $q\in[10^{-12},1-10^{-12}]$,
\begin{equation}
G_s=\frac1{N_s}\sum_{i=1}^{N_s}\!\left[\operatorname{LL}(y^{\mathrm{flip}}_{si},q^A_{si})-\operatorname{LL}(y^{\mathrm{flip}}_{si},q^B_{si})\right],\quad
\operatorname{LL}(y,q)=y\log q+(1-y)\log(1-q).
\label{eq:d-flip-loglik-gain}
\end{equation}
$G_s$ is augmented-minus-baseline gain in nats per donor row. Table~\ref{tab:d-flip-prediction} averages paired changes equally across seeds 22--26. The coordinates improve prediction without group indicators and within WW; with indicators, AUPRC decreases. This tests their incremental value, not unique attribution to the scalar information statistic. Original inputs, exact-source refits and paired records are in the artifact.

\begin{table}[htbp]
\centering\small
\setlength{\tabcolsep}{3pt}
\caption{Incremental flip prediction. Counts and rates are seed means; parentheses count seeds with improvement. Changes are paired within seed.}
\label{tab:d-flip-prediction}
\begin{tabular}{@{}p{.24\linewidth}rrccc@{}}
\toprule
Analysis & Rows & \shortstack{Flip\\rate} & \shortstack{Log-likelihood\\gain} & $\Delta$AUPRC & $\Delta$AUROC \\
\midrule
All groups, no indicators & 9,398 & 0.199 & $+0.016$ & $+0.035$ (5/5) & $+0.096$ (5/5) \\
All groups, with indicators & 9,398 & 0.199 & $+0.001$ & $-0.007$ (0/5) & $+0.003$ (5/5) \\
Selected slice & 1,181 & 0.564 & $+0.013$ & $+0.026$ (5/5) & $+0.015$ (5/5) \\
\bottomrule
\end{tabular}
\par\smallskip
\end{table}

\FloatBarrier
\subsection{Synthetic Classification and Tail Checks}
\label{app:synthetic-checks}

Draw $U_e\sim\mathrm{Uniform}(0,1)$, set $\delta_e=0.35U_e^{1/\gamma}$, and draw independent standard normals $s,r$ for each observation. The task is $z=s+\delta_er$, $Y=\mathbf1\{z>0\}$, scored by nuisance-only logit $4s$. Thus
\begin{equation}
\Pr(\delta_e<t)=(t/0.35)^\gamma\ (0\le t\le0.35),\qquad J_{TT,e}=1+\delta_e^2,\quad F_e^0=\delta_e^2.
\label{eq:d-synthetic-generator}
\end{equation}
Population feature moments are exact; loss/accuracy use 256 observations per environment. The $\gamma=1$ experiment uses 300,000 environments and seeds 11--15. Fit the log-survival slope over the largest 5\% of inverse information; empirical CVaR averages values at or above the empirical $\alpha$-quantile. Figure~\ref{fig:geometry-tails} regenerates seed 11 with $L_e=(\delta_e^2+10^{-12})^{-1}$, rather than recovering missing original raw output; its survival curve and the five-seed summaries have different replication levels.

Table~\ref{tab:d-synthetic-summary} checks Theorem~\ref{thm:inverse-information-geometry} and the scalar portfolio law in Appendix~\ref{app:theory-tails}. For $K$ independent draws, information $K^{-1}\sum_j\delta_{e,j}^2$ has exponent $K/2$ at $\gamma=1$; weights affect its constant. The mass sweep predicts $\gamma/2$. An anchor with $\delta_0=0.5$ and reserve mass $0.05$ yields information $0.95\delta_e^2+0.05(0.25)$, floor $0.0125$, and measured certificate $0.01260\pm0.000001$ (five seeds; Appendix~\ref{app:information-reserves}).

\begin{table}[htbp]
\centering\small
\setlength{\tabcolsep}{3pt}
\caption{Synthetic checks. Panel (a) and the branch rows are five-seed means $\pm$ nominal 95\% normal-approximation half-widths; zero widths reflect rounding. Mass-sweep rows are reported single-seed values.}
\label{tab:d-synthetic-summary}
\label{tab:d-structural-comparisons}
\textit{(a) Prediction and information}\par\smallskip
\begin{tabular}{@{}lc@{}}
\toprule
Quantity & Estimate \\
\midrule
Mean accuracy & $94.54\pm0.00\%$ \\
Fifth-percentile environment accuracy & $88.67\pm0.00\%$ \\
$\CVaR_{0.95}$ of environment-mean NLL & $0.262\pm0.000$ \\
$I_{0.95}^{\FG}$ & $(1.22\pm0.53)\times10^{-8}$ \\
Inverse-information exponent $\widehat\beta$ & $0.498\pm0.003$ \\
\bottomrule
\end{tabular}
\par\smallskip
\textit{(b) Inverse-information tail exponents}\par\smallskip
\begin{tabular}{@{}lccc@{}}
\toprule
Construction & Setting & Predicted $\beta$ & Estimated $\beta$ \\
\midrule
Mass near the null & $\gamma=1$ & 0.5 & 0.509 \\
Mass near the null & $\gamma=2$ & 1 & 0.985 \\
Mass near the null & $\gamma=4$ & 2 & 2.001 \\
Independent branches & $K=1$ & 0.5 & $0.499\pm0.003$ \\
Independent branches & $K=2$ & 1 & $1.000\pm0.009$ \\
Independent branches & $K=3$ & 1.5 & $1.508\pm0.015$ \\
\bottomrule
\end{tabular}
\par\smallskip
\end{table}

\FloatBarrier
\subsection{Hidden-Slice Discovery Results}
\label{app:discovery-results}

Appendix~\ref{app:hidden-slice-discovery} gives the blind protocol: task labels, frozen MobileNet features and a fixed CLIP direction are available; background/group metadata remain sealed through held-out evaluation. Results use the primary hidden-slice cohort, seeds 17--21, with full-validation gates frozen before test. Support correction may depend on the test sample but leaves all five primary CLIP gates' test weights unchanged. Information uses same-class pooling; accuracy, uncalibrated NLL and composition use admissible within-class weights without pooling. The artifact identifies this cohort as v2 and records original hashes; the older uncorrected cohort is excluded.

All selectors share classifiers with average accuracy $76.3\pm0.7\%$ and post hoc worst-group accuracy $57.3\pm1.3\%$. CLIP search finds $98.8\%$ membership-weighted slice accuracy. Table~\ref{tab:d-hidden-selectors} includes every matched-mass comparator and paired contrast. The random-direction interval and the uncentered and centered feature-moment intervals against high-loss selection contain zero. Feature controls reevaluate gates selected using predictive Fisher without rediscovery.

\begin{table}[H]
\centering\small
\setlength{\tabcolsep}{3pt}
\caption{Hidden-slice comparisons. Panel (a): means $\pm$ 95\% Student-$t$ half-widths across five paired seeds; predictive information retains calibration and shrinkage $\eta=0.10$. Panel (b): CLIP-search minus comparator, with paired 95\% Student-$t$ intervals; negative differences indicate lower information.}
\label{tab:d-hidden-selectors}
\label{tab:d-hidden-contrasts}
\textit{(a) Held-out selectors}\par\smallskip
\begin{tabular}{@{}p{.53\linewidth}cc@{}}
\toprule
Selector & \shortstack{Predictive\\projected information} & \shortstack{Uncalibrated\\slice NLL} \\
\midrule
Fisher search, fixed CLIP direction & $0.005\pm0.001$ & $0.031\pm0.017$ \\
High loss, same target within-class mass & $0.019\pm0.006$ & $5.779\pm0.496$ \\
Random, same target within-class mass & $0.013\pm0.001$ & $1.025\pm0.059$ \\
Constant class-conditional gate & $0.013\pm0.001$ & $1.027\pm0.049$ \\
Fisher search, random direction & $0.006\pm0.001$ & $0.031\pm0.015$ \\
\bottomrule
\end{tabular}
\par\smallskip
\textit{(b) Paired information contrasts}\par\smallskip
\begin{tabular}{@{}llcc@{}}
\toprule
Information & Comparison & Difference & 95\% interval \\
\midrule
Predictive & High-loss slice & $-0.014$ & $[-0.020,-0.008]$ \\
Predictive & Random slice & $-0.008$ & $[-0.009,-0.007]$ \\
Predictive & Constant gate & $-0.008$ & $[-0.009,-0.007]$ \\
Predictive & Random direction & $-0.0004$ & $[-0.0015,0.0007]$ \\
Feature moments & High-loss slice & $-0.034$ & $[-0.106,0.039]$ \\
Centered covariance & High-loss slice & $-0.071$ & $[-0.166,0.024]$ \\
\bottomrule
\end{tabular}
\par\smallskip
\end{table}

\paragraph{Class selection and overlap.}
CLIP, random-direction, random-slice and constant-gate selectors choose landbirds in all five seeds; high loss chooses landbirds in seeds 17,18,21 and waterbirds in 19,20. The two different-class Jaccards are zero; same-class values are $0.00222$, $0.00111$, and $0$. Their mean $0.000666$, reported as $0.001$, combines between-class and within-class differences under Appendix~\ref{app:discovery-crossfit}'s top-count rule.

\paragraph{Support and sensitivity.}
On same-class rows from the 15\% training-calibration subset, cross-fold gate-score correlation is $0.419\,[0.353,0.485]$ and overlap $0.258\,[0.225,0.291]$. Test ESS is $414\,[383,445]$ using admissible weights. Raw membership mass is $0.083\,[0.077,0.089]$ within the selected class (target $0.10$), not the full test set; landbirds contain 4,510 test examples. Validation cross-fitting, separate from final testing, gives information $0.004$--$0.005$ over masses $\{0.05,0.10,0.20\}$ versus high-loss $0.012$--$0.027$, and $0.004$--$0.006$ over shrinkages $\{0.05,0.10,0.20\}$.

Composition after held-out evaluation is $97.0\%$ landbird/land for CLIP, $94.4\%$ for random direction, and $58.4\%$ landbird/water for high loss. The hidden slice has lower projected predictive information than the weakest predefined group in all five seeds: mean paired difference $-0.0029$, with 95\% Student-$t$ interval $[-0.0034,-0.0024]$, taking the predefined-group minimum separately within each seed. Both sides use the same model, CLIP coordinate, calibrated predictive moments, pooling $\eta=0.10$, and ridge $10^{-3}$. A metadata-trained oracle probe used only after blind freeze gives information $0.003\pm0.000$, NLL $0.009\pm0.004$, and $99.1\%$ landbird/land composition (zero width is rounding). Repair outcomes are in Appendix~\ref{app:extended-training}.

\FloatBarrier
\subsection{Predefined-Group Diagnostics}
\label{app:predefined-diagnostics}

\paragraph{Fitting.}
CivilComments-WILDS~\citep{borkan2019nuanced,koh2021wilds} predicts comment toxicity, with $y=0$ denoting nontoxic comments. Identity-any indicates that at least one of the 24 original identity annotation scores is at least $0.5$. Training-fitted unigram/bigram TF-IDF (50,000 features, minimum document frequency 5, sublinear TF) feeds class-balanced SGD logistic task/identity-any probes ($\ell_2=10^{-5}$, 50 passes, tolerance $10^{-3}$). Identity-by-label groups require 100 examples; tail regression uses the largest 50\% of group inverse information, requiring eight groups and four tail points. Seed 13 gives the fixed diagnostic; seeds 13--17 give aggregates.

Waterbirds uses frozen ImageNet MobileNetV3-small features (extraction batch 64) and balanced L-BFGS logistic task/background models (1,000 maximum iterations). Four groups are reported directly, without fitting a tail exponent; only controls/anchors vary over seeds 17--21. Primary task/probe and metadata coordinates use training moments. These original runs used an i9-11900F/32 GB workstation, about one minute for CivilComments and three minutes for first Waterbirds extraction/fitting, then cached features.

\begin{table}[htbp]
\centering\small
\setlength{\tabcolsep}{3pt}
\caption{Original predefined-group diagnostics, distinct from fresh audits in Appendix~\ref{app:implemented-audit}. Means $\pm$ nominal 95\% normal-approximation half-widths; Waterbirds zeros reflect fixed-model fitting.}
\label{tab:d-predefined-diagnostics}
\begin{tabular}{@{}lcccc@{}}
\toprule
Dataset & \shortstack{Average\\accuracy (\%)} & \shortstack{Worst-group\\accuracy (\%)} & \shortstack{Loss-\\$\CVaR_{0.95}$} & $I_{0.95}^{\FG}$ \\
\midrule
CivilComments & $86.10\pm0.22$ & $43.75\pm0.64$ & $0.998\pm0.012$ & $0.500\pm0.001$ \\
Waterbirds & $77.11\pm0.00$ & $56.54\pm0.00$ & $1.741\pm0.000$ & $0.165\pm0.000$ \\
\bottomrule
\end{tabular}
\par\smallskip
\end{table}

For fixed CivilComments, the weakest-information group is \texttt{muslim}/$y=0$ (accuracy $60.1\%$); lowest accuracy is \texttt{black}/$y=0$ ($44.0\%$). Waterbirds' weakest-information group is landbird/land ($97.3\%$ accuracy), whereas waterbird/land has lowest accuracy ($56.5\%$); inverse-information p95 is $5.90$.

\paragraph{Specification and stress controls.}
The prespecified semantic family is $\{b_{\rm meta},b_{\rm probe}\}$: standardized identity-any/background annotation and its training-fitted probe, without cross-fitting. Semantic relevance follows those annotations; no held-out relevance threshold, anchor task-preservation tolerance, fold-stability admission screen, or nuisance-conditioning threshold was applied. The ridge is $10^{-3}$; family minima are $0.499$ and $0.165$. Reserve anchors (identity-present label-balanced CivilComments; four-group-balanced Waterbirds) are construction inputs, not admission samples. Fresh exact/ridge audits and hidden-gate stability are separate analyses.

\begin{table}[htbp]
\centering\small
\setlength{\tabcolsep}{3pt}
\caption{Nuisance sensitivity of fixed-model diagnostics. Entries are $I_{0.95}^{\FG}$; worst-group accuracy remains $44.0\%$/$56.5\%$ for CivilComments/Waterbirds.}
\label{tab:d-nuisance-sensitivity}
\begin{tabular}{@{}lccp{.27\linewidth}@{}}
\toprule
Specification & CivilComments & Waterbirds & Role \\
\midrule
Metadata & 0.587 & 0.266 & Semantic-family member \\
Learned probe & 0.499 & 0.165 & Semantic-family minimum \\
Random & 0.588 & 0.482 & Negative control \\
Under-projected & 0.588 & 0.482 & Under-projection control \\
Over-projected & 0.001 & 0.012 & Task-erasure control \\
Hidden-loss & 0.060 & 0.177 & Loss-coordinate stress control \\
\bottomrule
\end{tabular}
\par\smallskip
\end{table}

Random nuisance is independent standard normal; under-projection sets it to zero. Over-projection uses standardized task score plus Gaussian noise of standard deviation $0.01$ and deliberately erases task information. ``Hidden-loss'' is observed test loss with original groups fixed. The random, over-projected, and observed-loss coordinates use evaluated-array normalization and are not independently learned nuisance scores. The metadata/random contrast is $0.587$/$0.588$ on CivilComments, versus probe/random $0.165$/$0.482$ on Waterbirds, illustrating specification dependence.

The versioned artifact (see the \hyperref[sec:artifact-availability]{artifact availability statement}) retains execution commands, full-precision matrices and overlaps, run-history ledgers and expanded source tables, with original cohort provenance. Those records supplement the scientific definitions and controls above.
\FloatBarrier

\FloatBarrier

\clearpage
\section{Information-Changing Interventions}
\label{app:information-interventions}
This appendix gives the reserve proof, training and acquisition comparisons, and the influence derivative behind Section~\ref{sec:improving-information}. Distinct cohorts and information targets remain separate. Expanded protocols and numerical breakdowns are retained in the versioned artifact's \path{docs/extended_appendix_details/}, with executable reproduction guides and provenance (see the \hyperref[sec:artifact-availability]{artifact availability statement}).

\subsection{Information Reserves under Observed-Source Allocation}
\label{app:information-reserves}
\label{sec:information-floors}

Fix the predictor, coordinates, and normalization. Let $J_e,J_0\succeq0$ have positive-definite nuisance blocks. The anchor is an additional observation source with $F_0^0:=S_0(J_0)\succeq\rho_0I$, $\rho_0>0$. For the allocation in Section~\ref{sec:improving-information}, each observation's source is observed and allocation probabilities are parameter-independent. Sampling the anchor with probability $r\in(0,1]$ therefore gives
\begin{equation}
\begin{aligned}
J_e^{\res}&=(1-r)J_e+rJ_0,\\
S_0(J_e^{\res})&\succeq(1-r)S_0(J_e)+rS_0(J_0)\succeq r\rho_0I.
\end{aligned}
\label{eq:reserve-information-floor}
\end{equation}
The same statement applies to affine allocation of a fixed operational matrix map; it is not an assertion about an unobserved mixture likelihood.

\begin{proof}
For $J=\begin{psmallmatrix}A&B\\B^\top&C\end{psmallmatrix}$, $C\succ0$, completing the square gives
\begin{equation}
u^\top S_0(J)u=\min_v q_J(u,v),\qquad
q_J(u,v)=\begin{pmatrix}u\\-v\end{pmatrix}^{\!\top}J\begin{pmatrix}u\\-v\end{pmatrix},
\quad v_* = C^{-1}B^\top u.
\label{eq:reserve-schur-variational}
\end{equation}
Linearity in $J$ implies
\begin{equation}
\begin{aligned}
\min_v\{(1-r)q_{J_e}(u,v)+rq_{J_0}(u,v)\}
&\ge(1-r)\min_vq_{J_e}(u,v)+r\min_vq_{J_0}(u,v)\\
&\ge r\rho_0\lVert u\rVert^2.
\end{aligned}
\label{eq:reserve-schur-concavity}
\end{equation}
The last inequality uses $S_0(J_e)\succeq0$ and the anchor bound; holding for every $u$, it proves~\eqref{eq:reserve-information-floor}.
\end{proof}

For any environment distribution, $0<\alpha<1$ and $\varepsilon\ge0$, CVaR monotonicity gives
\begin{equation}
L_e^{\res,(\varepsilon)}\le(r\rho_0+\varepsilon)^{-1},
\qquad I_\alpha^{\FG,\res}(\varepsilon)\ge r\rho_0+\varepsilon\ge r\rho_0.
\label{eq:reserve-certificate-bound}
\end{equation}
Here $r=m/(n+m)$ for $m$ anchor and $n$ base observations. An estimated anchor's positive lower bound from Appendix~\ref{app:lower-audit} yields the same guarantee on its confidence event. The claim concerns the fixed predictor; retraining requires a new audit. Appendix~\ref{app:synthetic-checks} evaluates $r=.05$, $\rho_0=.25$, and the resulting floor $.0125$.

\subsection{Regularized Training}
\label{app:regularized-training}

For a declared group matrix $\widehat J_g(w)$, put $\widehat L_g(w)=[\lambda_{\min}(S_{\rid}(\widehat J_g(w)))+\varepsilon]^{-1}$. With fixed $\pi_g\ge0$, $\sum_g\pi_g=1$, training solves
\begin{equation}
\min_{w,\tau}\ \widehat R_{\mathrm{base}}(w)+\lambda_{\FG}
\left[\tau+\frac{1}{1-\alpha}\sum_g\pi_g(\widehat L_g(w)-\tau)_+\right],
\label{eq:fisher-glass-training-objective}
\end{equation}
Here $w$ denotes predictor parameters, $\widehat R_{\mathrm{base}}$ is the ERM or Group DRO task-risk objective, and $\lambda_{\FG}$ its information-penalty coefficient. Frozen-feature Waterbirds uses uncentered second moments of standardized task logits and fixed background-probe scores, ridge $10^{-3}$ and $\varepsilon=10^{-8}$. Training $\alpha=.75$ and audit $\alpha=.95$ both give the maximum inverse information over four equally weighted groups.

\paragraph{Protocol and selection.}
ImageNet MobileNetV3-small features and background-probe scores use training-moment standardization. A linear head receives 1,200 full-batch AdamW updates ($\mathrm{lr}=.01$, weight decay $10^{-4}$, seeds 17--21), Gaussian $N(0,.01^2)$ initial weights and zero intercept. Current logits use full-training population standard deviation plus $10^{-6}$; training projected information is floored at $\varepsilon$ before reciprocal stabilization. Evaluation uses fitted training-logit moments. Group DRO minimizes the maximum group mean cross-entropy; Loss-CVaR uses the largest $\lceil.1n\rceil$ training losses, and JTT refits with weight 20 on initial ERM mistakes.

For each FG family and seed, search $\Lambda_{\FG}=\{.01,.03,.1,.3\}$. Among candidates whose validation worst-group accuracy is within $.02$ of the family's best, maximize the validation certificate, breaking ties by higher validation worst-group accuracy. This uses validation group annotations. Historical choices are $.03$ for ERM + FG in all seeds and $.1$ or $.3$ for Group DRO + FG. Table~\ref{tab:e-frozen-performance} retains all six methods; official test data enter only final evaluation.

\begin{table}[htbp]
\centering\small
\caption{Original frozen-feature training cohort, seeds 17--21. Both panels give means $\pm$ nominal 95\% normal-approximation half-widths; accuracy is percent and the uncentered feature--Schur ridge certificate uses $\alpha=.95$.}
\label{tab:e-frozen-performance}\label{tab:e-frozen-information}
\begin{tabular}{@{}lrrr@{}}
\toprule
Method & Avg.\ acc.\ (\%) & Worst acc.\ (\%) & Loss-CVaR$_{.95}$\\
\midrule
ERM & $77.2\pm0.1$ & $55.7\pm0.3$ & $1.969\pm0.007$\\
Group DRO & $80.9\pm0.3$ & $67.3\pm0.3$ & $1.143\pm0.021$\\
JTT & $77.4\pm0.1$ & $54.1\pm0.2$ & $2.167\pm0.004$\\
Loss-CVaR & $78.2\pm0.1$ & $52.1\pm0.3$ & $2.427\pm0.041$\\
ERM + Fisher-Glass & $80.4\pm0.2$ & $56.3\pm0.3$ & $1.866\pm0.025$\\
Group DRO + FG & $80.6\pm0.9$ & $69.4\pm1.3$ & $0.855\pm0.082$\\
\midrule
Method & Ridge certificate & Inverse p95 & \\
\midrule
ERM & $0.198\pm0.001$ & $4.86\pm0.02$ & \\
Group DRO & $0.233\pm0.003$ & $4.10\pm0.06$ & \\
JTT & $0.234\pm0.001$ & $4.13\pm0.01$ & \\
Loss-CVaR & $0.233\pm0.002$ & $4.17\pm0.03$ & \\
ERM + Fisher-Glass & $0.399\pm0.006$ & $2.46\pm0.04$ & \\
Group DRO + FG & $0.614\pm0.059$ & $1.63\pm0.17$ & \\
\bottomrule
\end{tabular}
\end{table}

FG doubles ERM's certificate; adding it to Group DRO raises the certificate from $.233$ to $.614$ and mean worst-group accuracy from $67.3\%$ to $69.4\%$. These historical summaries differ from Appendix~\ref{app:estimation-certification}'s fresh ridge/exact audit cohort, which does not retroactively certify the historical checkpoints. Their original per-seed outputs were not recovered; retained reports, complete settings and independent rerun commands are in the artifact (see the \hyperref[sec:artifact-availability]{artifact availability statement}).

\subsection{Extended Training and Selection Results}
\label{app:extended-training}

\paragraph{End-to-end Group DRO + Fisher-Glass coefficient sweep.}
MobileNetV3-small is fine-tuned for 10 epochs (batch 128, evaluation batch 512, seeds 17--21) with AdamW learning rate and weight decay $10^{-4}$; the final epoch is evaluated. The reported H100 configuration uses bf16 and TF32. The nuisance is training-standardized background annotation; task logits use minibatch population standard deviation plus $10^{-6}$. Groups with fewer than two observations are omitted. Ridge is $10^{-3}$, $\alpha=.95$, and projected information is floored at $10^{-8}$ before another $10^{-8}$ enters the reciprocal. Table~\ref{tab:e-frontier-performance} retains the complete coefficient sweep: stronger penalties can increase information while degrading prediction.

\begin{table}[H]
\centering\small
\caption{End-to-end Waterbirds Group DRO + Fisher-Glass coefficient sweep. FG entries are means $\pm$ sample standard deviations over five seeds, not confidence intervals. The Group DRO reference is the baseline from the $\lambda_{\FG}=10^{-4}$ sweep; each sweep reran its baseline, so this is not a shared paired reference for every coefficient.}
\label{tab:e-frontier-performance}\label{tab:e-frontier-information}
\begin{tabular}{@{}lrrr@{}}
\toprule
Objective / $\lambda_{\FG}$ & Avg.\ acc.\ (\%) & Worst acc.\ (\%) & Loss-CVaR$_{.95}$\\
\midrule
Group DRO reference & $85.4$ & $53.2$ & $0.925$\\
$5\times10^{-5}$ & $79.7\pm6.2$ & $56.6\pm5.8$ & $0.733\pm0.061$\\
$10^{-4}$ & $79.5\pm10.2$ & $57.7\pm8.3$ & $0.733\pm0.110$\\
$3\times10^{-4}$ & $69.9\pm9.7$ & $58.8\pm8.7$ & $0.653\pm0.064$\\
$10^{-3}$ & $62.7\pm7.9$ & $49.9\pm1.8$ & $0.710\pm0.012$\\
$3\times10^{-3}$ & $57.7\pm4.5$ & $48.6\pm6.6$ & $0.717\pm0.036$\\
$10^{-2}$ & $47.9\pm16.5$ & $22.7\pm16.7$ & $1.031\pm0.416$\\
\midrule
Objective / $\lambda_{\FG}$ & Ridge certificate & Inverse p95 & \\
\midrule
Group DRO reference & $0.259$ & $3.68$ & \\
$5\times10^{-5}$ & $0.393\pm0.059$ & $2.49\pm0.34$ & \\
$10^{-4}$ & $0.502\pm0.166$ & $2.10\pm0.58$ & \\
$3\times10^{-4}$ & $0.614\pm0.189$ & $1.71\pm0.46$ & \\
$10^{-3}$ & $0.786\pm0.124$ & $1.29\pm0.20$ & \\
$3\times10^{-3}$ & $0.817\pm0.068$ & $1.22\pm0.10$ & \\
$10^{-2}$ & $0.809\pm0.076$ & $1.23\pm0.12$ & \\
\bottomrule
\end{tabular}
\end{table}

At $\lambda_{\FG}=10^{-4}$ the certificate rises from $.259$ to $.502$ and inverse p95 falls from $3.68$ to $2.10$; at $\lambda_{\FG}=10^{-2}$ worst-group accuracy is only $22.7\%$. Historical report summaries support this table; Appendix~\ref{app:estimation-certification}'s fresh audits are a separate cohort.

\paragraph{Prospective information training.}
Use Appendix~\ref{app:prospective-intervention}'s shared model, 50-epoch schedule and seeds 22--26. Backbone fine-tuning uses AdamW with learning rate and weight decay $10^{-4}$. FG adds coefficients $\{.003,.01,.03\}$ and a background probe trained on detached features (AdamW learning rate $.001$, default weight decay $.01$), whose scores are detached in the penalty. After one penalty-free epoch, training uses $\alpha=.75$, ridge $.001$, batch standard-deviation guard $10^{-6}$, groups of at least six observations with nuisance standard deviation at least $.05$, and a $.05$ information floor before direct inversion. This floor differs from audit stabilization.

For this prospective 50-epoch cohort, select the final model with largest validation certificate among candidates whose overall validation accuracy is at least ERM minus $.01$. If none qualify, maximize the certificate over all FG candidates for that seed and record the fallback. Choices are $[.01,.03,.01,.003,.01]$, with fallbacks in seeds 22 and 24. All audit quantities are frozen before intervention scoring. In the prospective replication, FG raises the validation certificate from $.289$ to $.330$, but the controlled drop is $52.9$ versus $52.6$ percentage points for ERM and the originally-correct flip rate is $57.7\%$ versus $56.4\%$. The accuracy constraint fails in two seeds and background sensitivity persists.

\paragraph{Constrained hidden-slice repair.}
The primary hidden-slice cohort of Appendix~\ref{app:discovery-results} compares loss, Fisher and combined repair without opening group metadata. Each arm adds its repair penalties to ERM, includes the unchanged model, and must preserve overall, balanced and each-class validation accuracy within $.02$ of the base. Loss coefficients are $\{.1,.3,1\}$, Fisher coefficients $\{.01,.03,.1\}$, and combined candidates use their Cartesian product. The same warm start and fresh AdamW optimizer ($\mathrm{lr}=.01$, weight decay $10^{-4}$, gradient clip 10) receive at most 1,200 steps. Calibration NLL at temperature one is checked every 25 steps, with patience 12 and improvement threshold $10^{-5}$; the best checkpoint is restored, with the initial model eligible for selection. The matched ERM continuation sets both penalties to zero. Base temperature remains fixed for information training, selection and subsequent audits.

For the frozen training slice, the dimensionless Fisher penalty is
\begin{equation}
P_{\mathrm{FG}}(w)=\frac{\max\{F_{\mathrm{init}},\varepsilon\}}{F(w)+\varepsilon},
\qquad \varepsilon=10^{-8},
\label{eq:e-dimensionless-penalty}
\end{equation}
Here $F(w)$ is the frozen slice's projected predictive information and $F_{\mathrm{init}}$ its value at the unchanged model, with shrinkage $.1$, ridge $.001$, and negative numerical information clipped to zero. Within the accuracy envelope, loss/Fisher repair minimizes frozen-slice loss/maximizes information. Combined selection uses the loss--information Pareto frontier, maximizing the smaller relative improvement; ties favor mean relative improvement, then the smaller coefficient sum normalized by grid maxima, $\lambda_{\mathrm{loss}}+10\lambda_{\mathrm{Fisher}}$, then earlier index. Fresh rediscovery uses matched gate/fold seeds, so the reported population can change.

\begin{table}[H]
\centering\small
\caption{Constrained repair on the primary hidden-slice cohort, seeds 17--21. Accuracy, NLL and information are means $\pm$ 95\% Student-$t$ half-widths; composition is a mean percentage. Panel two audits each arm's rediscovered CLIP-defined slice. The Group DRO oracle uses group annotations.}
\label{tab:e-repair-performance}\label{tab:e-repair-slices}
\begin{tabular}{@{}lrrr@{}}
\toprule
Arm & Avg.\ acc.\ (\%) & Worst acc.\ (\%) & Unchanged\\
\midrule
Matched ERM continuation & $76.3\pm0.7$ & $57.3\pm1.3$ & ---\\
Loss repair & $76.3\pm0.6$ & $56.9\pm1.3$ & $3/5$\\
Fisher repair & $76.3\pm0.7$ & $57.3\pm1.3$ & $5/5$\\
Combined repair & $76.4\pm0.9$ & $56.7\pm1.6$ & $4/5$\\
Oracle Group DRO & $80.0\pm0.3$ & $64.8\pm0.9$ & ---\\
\midrule
Arm & Slice NLL & Projected info. & LL mass (\%)\\
\midrule
Matched ERM continuation & $0.031\pm0.017$ & $0.005\pm0.001$ & $97.0$\\
Loss repair & $0.023\pm0.017$ & $0.005\pm0.002$ & $97.8$\\
Fisher repair & $0.031\pm0.017$ & $0.005\pm0.001$ & $97.0$\\
Combined repair & $0.031\pm0.017$ & $0.005\pm0.001$ & $97.2$\\
Oracle Group DRO & $0.033\pm0.018$ & $0.005\pm0.002$ & $95.9$\\
\bottomrule
\end{tabular}
\end{table}

Fisher repair chooses the unchanged model in every seed. Paired worst-group changes versus matched continuation are $-.4$, $.0$ and $-.6$ points for loss, Fisher and combined repair, with intervals containing zero. The group-supervised Group DRO oracle reaches $64.8\%$ worst-group accuracy while rediscovered information remains $.005$. LL denotes landbird/land.

\paragraph{Post hoc group-free coefficient selection.}
This analysis was designed after earlier group/test outcomes were inspected. Its grid $\{0,.01,.03,.1,.3\}$ uses~\eqref{eq:e-dimensionless-penalty}, not the known-group objective's scale. Training is partitioned $70/15/15\%$ for fitting/calibration/gate discovery; a frozen train-only slice supplies the penalty. All candidates share the warm start, training seed, optimizer, 1,200-step budget, checkpoint rule and base temperature above. Official validation enters only after heads are fixed; group metadata stays sealed through blind test evaluation.

Let $\widehat R_\lambda$ be mean fixed-temperature validation NLL, $s_\lambda=\operatorname{sd}_i(\ell_{\lambda i})/\sqrt{1199}$ its within-run standard error, and $\lambda_R=\arg\min_\lambda(\widehat R_\lambda,\lambda)$ lexicographically. Here $\ell_{\lambda i}$ is the fixed-temperature validation NLL of example $i$, and $\operatorname{sd}_i$ is the sample standard deviation with divisor $1198$. Five task-stratified validation folds rediscover candidate-specific slices; $V_{k,\lambda}$ is held-out inverse predictive information and $s_{V,\lambda}$ its sample standard deviation. Selection uses
\begin{equation}
\begin{aligned}
\Lambda_{\mathrm{risk}}&=\{\lambda:\widehat R_\lambda\le\widehat R_{\lambda_R}+s_{\lambda_R}\},&
U_\lambda^{\mathrm{adj}}&=\overline V_\lambda+t_{4,.99}s_{V,\lambda}/\sqrt5,\\
\widehat\lambda_{\mathrm{hidden}}
&=\min\arg\max_{\lambda\in\Lambda_{\mathrm{risk}}}(U_\lambda^{\mathrm{adj}})^{-1},&
t_{4,.99}&=3.74695 .
\end{aligned}
\end{equation}
The risk allowance uses the minimizer's SE for all candidates. Bonferroni adjustment covers five coefficients, but overlapping gate-training folds make this a ranking score without independent-fold coverage. Accuracy screens and paired inverse-information tests are diagnostics, not selection constraints. The group-supervised comparator maximizes validation worst-group accuracy over the same $\Lambda_{\mathrm{risk}}$, breaking ties toward smaller $\lambda$.

Matched zero continuation restores step zero in all five seeds, explaining equality with frozen ERM despite intervening optimization. Group-free selection equals plain NLL minimization in every seed; the information criterion does not change any choice. Versus matched continuation, paired mean changes are $+2.1$ average-accuracy points ($[+.2,+4.0]$), $+.006$ information ($[+.001,+.010]$), and $-12.6$ worst-group points ($[-21.6,-3.7]$). The group-supervised comparator opens validation annotations only after blind decisions and records are fixed and is not an upper bound. The post hoc coefficient-selection results and primary repair cohort have separate artifact identifiers (see the \hyperref[sec:artifact-availability]{artifact availability statement}), alongside extended settings and full tables.

\begin{table}[H]
\centering\small
\caption{Post hoc coefficient selection, five paired seeds. Accuracies are test results; information is projected predictive information in test slices defined by validation-rediscovered, frozen gates. Brackets are 95\% Student-$t$ intervals. Policies selecting identical models across these seeds share rows; the text explains both coincidences. Coefficient counts are $\lambda:\text{number of seeds}$.}
\label{tab:e-groupfree-performance}\label{tab:e-groupfree-information}
\begin{tabular}{@{}lrr@{}}
\toprule
Policy & Average accuracy (\%) & Worst accuracy (\%)\\
\midrule
Frozen ERM / matched zero & $75.5\ [74.3,76.6]$ & $55.7\ [53.3,58.2]$\\
Fixed $\lambda=.03$ & $76.6\ [76.1,77.2]$ & $51.5\ [47.1,55.8]$\\
Risk only / group-free & $77.5\ [76.5,78.6]$ & $43.1\ [33.1,53.1]$\\
Group-supervised comparator & $77.7\ [76.6,78.8]$ & $44.2\ [34.2,54.3]$\\
\midrule
Policy & Coefficient counts & Slice information\\
\midrule
Frozen ERM / matched zero & --- / $0:5$ & $0.007\ [0.005,0.009]$\\
Fixed $\lambda=.03$ & $.03:5$ & $0.007\ [0.004,0.010]$\\
Risk only / group-free & $0:1,\ .3:4$ & $0.012\ [0.006,0.019]$\\
Group-supervised comparator & $0:1,\ .1:1,\ .3:3$ & $0.011\ [0.005,0.018]$\\
\bottomrule
\end{tabular}
\end{table}

\paragraph{Operational choice of the penalty.}
For group-free selection, we propose prespecifying an information requirement and a validation-loss budget, then choosing the smallest coefficient satisfying both. If no candidate qualifies, report the requirement as unmet. With neither a specified information requirement nor a group-labeled audit set, retain $\lambda_{\FG}=0$. A worst-group-accuracy objective requires group labels or a justified group proxy. This threshold-based policy requires prospective validation; the reported experiment maximizes information within an NLL eligibility set.

\subsection{Tail-Transverse Influence}
\label{app:tail-transverse-influence}
\label{sec:targeted-interventions}

Fix the predictor, coordinates, calibration, slice rules and finite audit weights $\pi_g\ge0$, $\sum_g\pi_g=1$. An action specifies a positive-semidefinite path $J_g(t,a)=J_g+tH_{g,a}+o(t)$ as $t\downarrow0$, with symmetric $H_{g,a}$ and an operator-norm remainder. Allocation, normalization and retained pooling belong in this path; audit weights stay fixed.

\paragraph{One differential for exact and ridge targets.}
Write $J_g=\begin{psmallmatrix}A_g&B_g\\B_g^\top&C_g\end{psmallmatrix}$. Choose a fixed $\rho\ge0$ with $C_g+\rho I\succ0$: $\rho=0$ gives exact Schur information and $\rho=\lambda_{\rid}$ gives the ridge target. Set $Q_g=(C_g+\rho I)^{-1}$, $S_\rho(J_g)=A_g-B_gQ_gB_g^\top$, and let $\lambda_g$ be its smallest eigenvalue. Since $DQ_g[H_{NN}]=-Q_gH_{NN}Q_g$,
\begin{equation}
\begin{aligned}
DS_\rho(J_g)[H]={}&H_{TT}-H_{TN}Q_gB_g^\top-B_gQ_gH_{TN}^\top\\
&+B_gQ_gH_{NN}Q_gB_g^\top,\\
p_g(u)&=\begin{pmatrix}u\\-Q_gB_g^\top u\end{pmatrix},
\qquad u^\top DS_\rho(J_g)[H]u=p_g(u)^\top Hp_g(u).
\end{aligned}
\label{eq:influence-schur-differential}
\end{equation}
For a simple minimum eigenvalue with unit eigenvector $u_g$, differentiating the eigenvalue equation therefore gives
\begin{equation}
h_{g,a}:=D_a\lambda_g=p_g^\top H_{g,a}p_g,
\qquad p_g=p_g(u_g).
\label{eq:directional-schur-derivative}
\end{equation}
Let $s_a$ be the candidate's score vector and $M_{g,a}$ its matrix contribution to group $g$. For allocation rate $\eta_g\ge0$, additive and normalized contributions have different derivatives:
\begin{equation}
\begin{aligned}
H_{g,a}=\eta_gs_as_a^\top
&\ \Longrightarrow\ h_{g,a}=\eta_g(p_g^\top s_a)^2,\\
J_g(t,a)=\frac{J_g+t\eta_gM_{g,a}}{1+t\eta_g}
&\ \Longrightarrow\ H_{g,a}=\eta_g(M_{g,a}-J_g),\\
&\hspace{13mm}h_{g,a}=\eta_g\!\left[p_g^\top M_{g,a}p_g-\lambda_g
+\rho\lVert Q_gB_g^\top u_g\rVert^2\right].
\end{aligned}
\label{eq:influence-ridge-normalized-gain}
\end{equation}
Indeed $Q_gC_gQ_g=Q_g-\rho Q_g^2$, so $p_g^\top J_gp_g=\lambda_g-\rho\lVert Q_gB_g^\top u_g\rVert^2$; the correction vanishes at $\rho=0$. Per unit allocation, the additive rank-one contribution is the squared nuisance-adjusted residual $[u_g^\top(s_{T,a}-B_gQ_gs_{N,a})]^2$; at $\rho=0$, this is the efficient-score residual. Replacement uses the allocated new-minus-removed contribution. For one observation appended to group $g$, its indicator and original sample-count scaling enter $\eta_g$; they cannot be omitted when claiming a normalized-augmentation derivative.

\paragraph{Tail sensitivity, including nonsmooth cases.}
Fix $\varepsilon>0$ and $0<\alpha<1$, and put $L_g=(\lambda_g+\varepsilon)^{-1}$. Appendix~\ref{app:theory-certificate}'s finite CVaR dual uses
$\mathcal W_{\alpha,\pi}=\{\omega\ge0:\sum_g\omega_g=1,\ \omega_g\le\pi_g/(1-\alpha)\}$.
Let $\Omega_\alpha^\star(L)$ maximize $\sum_g\omega_gL_g$ over this set. If the maximizer $\omega_\alpha$ is unique and eigenvalues are simple,
\begin{equation}
G_\alpha(a):=-D_a\CVaR_{\alpha,\pi}(L)
=\sum_g\omega_{\alpha,g}\frac{h_{g,a}}{(\lambda_g+\varepsilon)^2},
\qquad D_a I_{\alpha,\pi}=I_{\alpha,\pi}^{\,2}G_\alpha(a).
\label{eq:tail-information-gain}
\end{equation}
Thus $G_\alpha(a)$ is the first-order decrease in inverse-information CVaR along action $a$'s specified allocation path. It follows from $D_aL_g=-h_{g,a}/(\lambda_g+\varepsilon)^2$; the dual weights include fractional threshold mass and upper-tail normalization.

For a repeated minimum eigenvalue with eigenspace $\mathcal E_g$, symmetric eigenvalue perturbation gives the first formula below. At a CVaR tie, the right derivative of a maximum is the maximum of its active directional derivatives; negation gives the second:
\begin{equation}
\begin{aligned}
h_{g,a}&=\min_{\substack{u\in\mathcal E_g\\\lVert u\rVert=1}}
p_g(u)^\top H_{g,a}p_g(u),\\
G_\alpha^+(a):=-D_a^+\CVaR_{\alpha,\pi}(L)
&=\min_{\omega\in\Omega_\alpha^\star(L)}
\sum_g\omega_g\frac{h_{g,a}}{(\lambda_g+\varepsilon)^2}.
\end{aligned}
\label{eq:influence-tied-cvar-gain}
\end{equation}
These formulas cover either or both sources of nondifferentiability; the reciprocal-certificate identity holds with right derivatives. An arbitrary fixed maximizing tail vector is only a score unless it attains this direction-dependent minimum.

For assigned label $y_a$, tail-transverse influence multiplies the directional information gain by current prediction loss:
\begin{equation}
\TTI(a)=G_\alpha(a)\loss_w(x_a^{\mathrm{cf}},y_a),
\qquad \loss_w(x,y)=-\log p_w(y\mid x),
\label{eq:tail-transverse-influence}
\end{equation}
using $G_\alpha^+$ at nonsmooth points. This ranks actions at the fixed predictor; finite augmentation, refitting and changed normalization require a new audit. The implemented residual-square surrogate $\widetilde G_\alpha$ below omits allocation/baseline terms and is distinguished from this derivative.

\subsection{Feature-Space Acquisition Experiments}
\label{app:feature-acquisition}

\paragraph{Candidates, budgets and refitting.}
Using training-standardized frozen MobileNetV3-small representations, let $\bar v_{y,b}$ denote the official-training mean for task label $y$ and background $b\in\{0,1\}$. Each official-validation candidate becomes
\begin{equation}
v^{\mathrm{cf}}=v+\gamma_{\mathrm{cf}}(\bar v_{y,1-b}-\bar v_{y,b}),
\qquad y^{\mathrm{cf}}=y,\quad b^{\mathrm{cf}}=1-b .
\label{eq:feature-counterfactual}
\end{equation}
This preserves the assigned label, not necessarily semantic or pixel-level content. The reported unit-strength run uses label-conditional means for every candidate. A balanced logistic background probe, fitted and standardized on training data, stays fixed and scores the translated features. Test data supply final evaluation only.

Budgets $b_{\mathrm{acq}}=.01,.02,.05$ use $k=\min\{\max(1,\operatorname{round}(b_{\mathrm{acq}}n_{\mathrm{train}})),n_{\mathrm{cand}}\}$, giving $48,96,240$ candidates from the 1,199-example validation pool relative to 4,795 original training examples. These are formula-implied counts; historical selected-index records were not recovered. Acquired features and labels augment training, and a balanced logistic task head is fitted from scratch, without an FG penalty or special acquisition weights. Both probe and head use L-BFGS, at most 1,000 iterations, and library defaults ($C=1$, tolerance $10^{-4}$, fitted intercept). Class weights are recomputed on the augmented set. New task logits are normalized by augmented-training moments, while nuisance normalization stays fixed. Acquired validation examples are no longer an independent evaluation sample. Seeds are 17--21; the historical scikit-learn version was not pinned.

\paragraph{Implemented score and its scope.}
Let $(z,q)$ be the task logit and nuisance score standardized on original training data. Original validation-group moments are unweighted second moments, without $p(1-p)$; put $\beta_g=J_{TN,g}/(J_{NN,g}+.001)$, $\lambda_g=J_{TT,g}-J_{TN,g}^2/(J_{NN,g}+.001)$, and $L_g=(\lambda_g+\varepsilon)^{-1}$ with $\varepsilon=10^{-8}$. Numerical negative information is floored at zero. For the threshold-inclusive tail $T=\{g:L_g\ge\operatorname{quantile}_{.95}(L_1,\ldots,L_4)\}$, the code ranks by
\begin{equation}
\widetilde G_\alpha(a)=
\frac{\sum_{g\in T}(\lambda_g+\varepsilon)^{-2}(z_a^{\mathrm{cf}}-\beta_gq_a^{\mathrm{cf}})^2}
{\max\{\sum_{g\in T}(\lambda_g+\varepsilon)^{-2},10^{-12}\}},
\qquad
\widetilde{\TTI}_\alpha(a)=\widetilde G_\alpha(a)\loss_w(v_a^{\mathrm{cf}},y_a).
\label{eq:implemented-acquisition-score}
\end{equation}
Every tail projector is applied to every candidate, irrespective of its assigned group; for a unique worst group the inverse-square weight cancels. The score lacks group allocation, sample-count scaling, old-information subtraction and replacement terms. Its nonnegative residual square is an unnormalized rank-one contribution at fixed coordinates, not the derivative of normalized augmentation or of the post-refit certificate. Threshold-inclusive ties also differ from the fractional/direction-dependent CVaR convention in Appendix~\ref{app:tail-transverse-influence}.

High loss uses $\loss_w$; pure tail selects by $\widetilde G_\alpha$ alone. Rank product maps ascending \texttt{argsort} positions for each score evenly from 0 to 1, then multiplies their rank scores; equal values receive separate ranks. Uncertainty product multiplies $\widetilde G_\alpha$ by $p(1-p)$. The acquisition baseline has $78.2\%$ average and $52.5\%$ worst-group accuracy, certificate $.209$, and weakest-information group landbird/land; it is distinct from the predefined-diagnostic cohort.

\begin{table}[htbp]
\centering\small
\caption{Feature-space acquisition, means over seeds 17--21. $\Delta a$ is worst-group accuracy gain in percentage points and $\Delta I$ the absolute ridge-certificate gain. Panel (a) gives the complete strength-by-budget comparison; panel (b) retains pure-tail and alternative-product controls at unit strength. Only aggregate reports survive, so historical seed-level intervals are unavailable.}
\label{tab:acquisition-strength}\label{tab:acquisition-budget}\label{tab:acquisition-products}
\begin{tabular}{@{}llrrrr@{}}
\toprule
\multicolumn{6}{l}{(a) Strength and budget: implemented geometric score $\times$ loss versus high loss}\\
 & & \multicolumn{2}{c}{$\widetilde{\TTI}_\alpha$} & \multicolumn{2}{c}{High loss}\\
\cmidrule(lr){3-4}\cmidrule(l){5-6}
$\gamma_{\mathrm{cf}}$ & Budget & $\Delta a$ & $\Delta I$ & $\Delta a$ & $\Delta I$\\
\midrule
0.25 & $1\%$ & $+4.05$ & $+0.0308$ & $+5.76$ & $+0.0411$\\
 & $2\%$ & $+7.17$ & $+0.0366$ & $+7.63$ & $+0.0477$\\
 & $5\%$ & $+8.88$ & $+0.0521$ & $+10.12$ & $+0.0590$\\
0.5 & $1\%$ & $+6.39$ & $+0.0205$ & $+4.36$ & $+0.0328$\\
 & $2\%$ & $+5.61$ & $+0.0300$ & $+4.83$ & $+0.0446$\\
 & $5\%$ & $+6.54$ & $+0.0417$ & $+8.57$ & $+0.0437$\\
1 & $1\%$ & $+4.52$ & $+0.0317$ & $+3.27$ & $+0.0603$\\
 & $2\%$ & $+6.85$ & $+0.0452$ & $+5.76$ & $+0.0581$\\
 & $5\%$ & $+7.94$ & $+0.0606$ & $+7.01$ & $+0.0630$\\
1.5 & $1\%$ & $+7.48$ & $+0.0662$ & $+7.63$ & $+0.1018$\\
 & $2\%$ & $+8.72$ & $+0.0826$ & $+9.81$ & $+0.1006$\\
 & $5\%$ & $+9.97$ & $+0.1159$ & $+12.31$ & $+0.1137$\\
\bottomrule
\end{tabular}

\smallskip
\begin{tabular}{@{}lrrrrrr@{}}
\toprule
\multicolumn{7}{l}{(b) Unit-strength controls}\\
 & \multicolumn{2}{c}{Pure tail} & \multicolumn{2}{c}{Rank product} & \multicolumn{2}{c}{Score $\times p(1-p)$}\\
\cmidrule(lr){2-3}\cmidrule(lr){4-5}\cmidrule(l){6-7}
Budget & $\Delta a$ & $\Delta I$ & $\Delta a$ & $\Delta I$ & $\Delta a$ & $\Delta I$\\
\midrule
$1\%$ & $+0.62$ & $+0.0045$ & $+5.45$ & $+0.0287$ & $+1.56$ & $+0.0012$\\
$2\%$ & $+2.80$ & $+0.0098$ & $+6.54$ & $+0.0326$ & $+1.71$ & $+0.0073$\\
$5\%$ & $+3.58$ & $+0.0340$ & $+6.07$ & $+0.0516$ & $+3.12$ & $+0.0230$\\
\bottomrule
\end{tabular}
\end{table}

At $\gamma_{\mathrm{cf}}=1$, the loss product exceeds high loss in mean worst-group accuracy at all budgets, while high loss gives larger certificate gains. This advantage is strength-dependent: high loss wins accuracy at every budget for $.25$ and $1.5$, and at the $5\%$ budget for $.5$. The rank product also exceeds the chosen loss product at $1\%$ ($5.45$ versus $4.52$ points); the chosen product is not uniformly best. Information gains and predictive gains are distinct outcomes.

Table~\ref{tab:acquisition-strength} reproduces historical reports, not fresh runs. Missing per-seed JSON/CSV, selected indices and pinned versions preclude uncertainty reconstruction. The \hyperref[sec:artifact-availability]{artifact} supplies rerun code, aggregate candidate score/loss summaries, allocation mechanics and protocols. Five seeds on fixed splits do not constitute five independent datasets.
\FloatBarrier

\FloatBarrier

\clearpage
\section*{Artifact Availability}
\label{sec:artifact-availability}

\paragraph{Code and supporting artifacts.}
The public repository and versioned release \texttt{camera-ready-v22} are available at
\begin{center}
\url{https://github.com/E-zClap/fisher-glass-learning/releases/tag/camera-ready-v22}.
\end{center}
The release provides experiment code, dependencies, derived numerical inputs, per-seed records, checksums, and commands for reproducing the retained analyses and independently rerunning the experiments. Its reproduction guides specify data preparation, environments, and experiment settings. Its artifact manifest maps manuscript tables to the supporting files and identifies the corrected hidden-slice cohort. Expanded appendix sources retain displaced execution details and numerical records; the manuscript map records consolidated table labels. The prospective matrix and auxiliary flip-prediction reanalyses can be replayed from the included frozen numerical inputs. Selected image examples used in Figure~\ref{fig:concept} are included with source provenance. Full benchmark datasets and pretrained weights are not redistributed; dataset sources and access terms are documented separately. Historical synthetic and frozen-training summaries and the acquisition results retain their stated provenance: where original per-seed outputs are unavailable, the release includes the supporting aggregate reports and executable rerun workflows; fresh experiment replication does not imply exact replay of missing historical outputs.

\end{document}